\pdfoutput=1

\documentclass[11pt,table,xcdraw]{article}
\usepackage[final]{acl}

\usepackage{times}

\usepackage{latexsym}
\usepackage{amssymb} 
\usepackage{booktabs, tabularx, makecell, placeins}
\usepackage{etoolbox}   
\usepackage{array}
\usepackage{multirow}
\usepackage{pifont}
\usepackage{booktabs}
\usepackage{siunitx}
\usepackage{array}
\usepackage{geometry}
\usepackage{enumitem}
\setlist[itemize]{left=0.25em, topsep=1pt, itemsep=1pt}
\usepackage[htt]{hyphenat} 
\usepackage{subcaption}
\usepackage[export]{adjustbox}

\usepackage{pifont}
\newcommand{\yes}{\ding{51}}   
\newcommand{\no}{\ding{55}}    

\newcommand{\rowcolours}{%
  \ifnum\value{rownum}=1\relax\else
    \ifodd\value{rownum}\relax\rowcolor{oddrow}\else\rowcolor{evenrow}\fi
  \fi}
\newcolumntype{Y}{>{\RaggedRight\arraybackslash}X}

\definecolor{rowgray}{gray}{0.95}

\usepackage{tabularx,makecell}
\definecolor{hdrblue}{HTML}{D5E4FA}   
\definecolor{oddrow}{HTML}{F6F8FC}    
\definecolor{evenrow}{HTML}{EDF3FE}   

\definecolor{lightrow}{HTML}{F5F5F5}
\definecolor{headerblue}{HTML}{D9E2F3}
\usepackage[T1]{fontenc}
\usepackage{tabularray}
\usepackage{float}
\usepackage{ragged2e}
\usepackage{graphicx}

\NewColumnType{Y}{>{\RaggedRight\arraybackslash}X}

\definecolor{dark}{HTML}{38598A}

\usepackage[utf8]{inputenc}

\usepackage{microtype}

\usepackage{inconsolata}

\definecolor{niceorange}{HTML}{F86624}

\newcommand{\dataset}{MedPath}
\newcommand{\biokg}{BioKG}
\newcommand{\biokgs}{BioKGs}

\newcommand{\cmark}{\ding{51}}%
\newcommand{\xmark}{\ding{55}}%

\title{\dataset{}: Multi-Domain Cross-Vocabulary Hierarchical Paths for Biomedical Entity Linking}

\author{
    Nishant Mishra\textsuperscript{1,2,}\thanks{\quad Corresponding author} \quad
    Wilker Aziz\textsuperscript{3}\quad
    Iacer Calixto\textsuperscript{1,2}\\
    \textsuperscript{1}Department of Medical Informatics, Amsterdam UMC, University of Amsterdam,\\ The Netherlands \\
    \textsuperscript{2}Amsterdam Public Health, Methodology, Amsterdam, The Netherlands \\
    \textsuperscript{3}ILLC, University of Amsterdam, The Netherlands \\
    \texttt{\{n.mishra, i.coimbra\}@amsterdamumc.nl} \quad \texttt{w.aziz@uva.nl}
}

\begin{document}
\maketitle
\begin{abstract}
Progress in biomedical Named Entity Recognition (NER) and Entity Linking (EL) is currently hindered by a fragmented data landscape, a lack of resources for building explainable models, and the limitations of semantically-blind evaluation metrics.
To address these challenges, we present \dataset{}, a large-scale and multi-domain biomedical EL dataset that builds upon nine existing expert-annotated EL datasets. In \dataset{}, all entities are 1) normalized using the latest version of the Unified Medical Language System (UMLS), 2) augmented with mappings to 62 other biomedical vocabularies and, crucially, 3) enriched with full ontological paths---i.e., from general to specific---in up to 11 biomedical vocabularies.
\dataset{} directly enables new research frontiers in biomedical NLP, facilitating training and evaluation of semantic-rich and interpretable EL systems, and the development of the next generation of interoperable and explainable clinical NLP models.
\end{abstract}

\section{Introduction}
\label{sec:intro}

\begin{figure*}[ht]
  \centering
  \includegraphics[width=\textwidth]{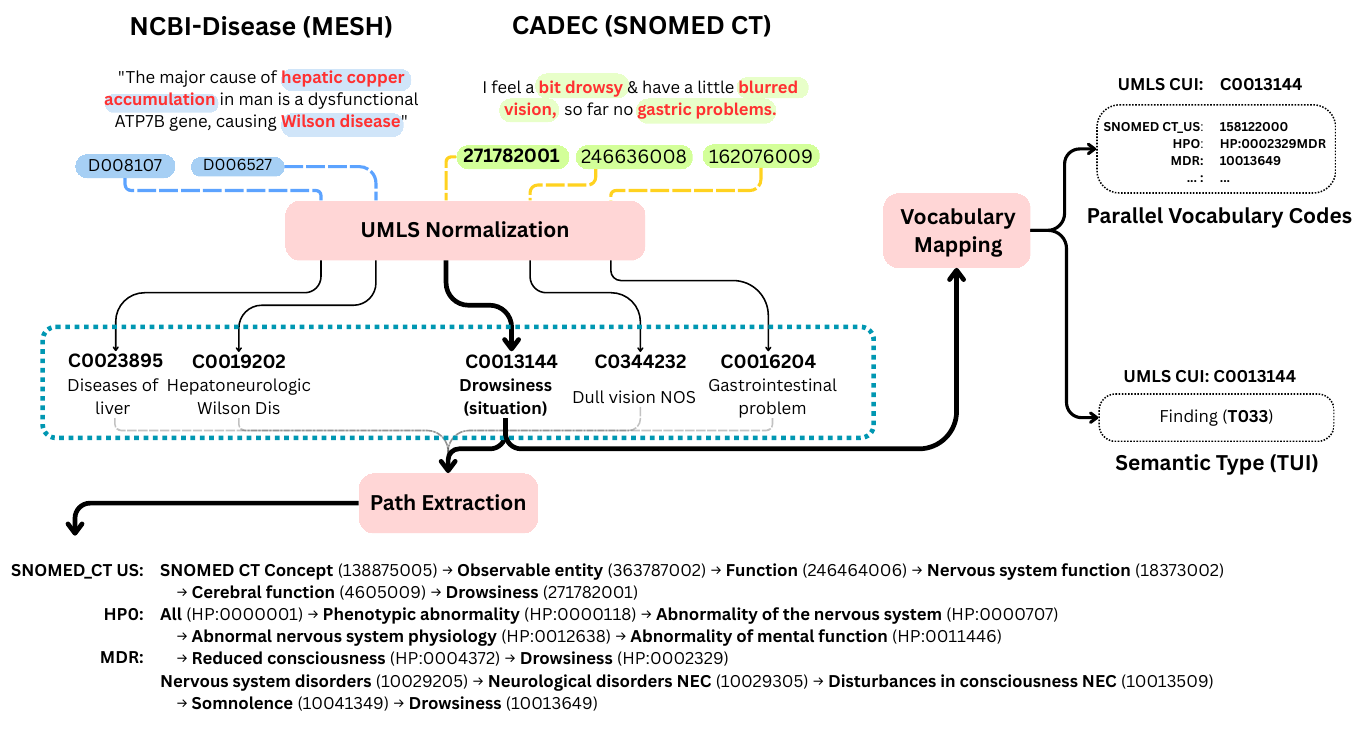}
  \caption{\dataset{} creation process. For illustration purposes, we show one example from two different datasets, and vocabulary mappings and path annotations for only one of the concepts, e.g., \textbf{C0013144 Drowsiness (situation).}}
  \label{fig:overview}
\end{figure*}
Health-related textual narratives abound in the healthcare domain and can be found for example in a patient's electronic health record~\cite[][]{Johnson2023}, in scientific research papers~\cite{Pubmed}, or in social media posts~\cite{basaldella-etal-2020-cometa}.
Making sense of and integrating the medical concepts in these narratives is a complex task that requires in-depth domain knowledge. 
Named Entity Recognition (NER) and Entity Linking (EL) are two foundational tasks in clinical NLP whose main goal is \textit{structuring the unstructured} \cite{roder2018gerbil}. In NER, we wish to identify all mention spans of clinically relevant entities in some input text (e.g., all drug mentions in a clinical progress note; \citealt{Nadeau2007ASO}).
In EL, we go a step further and wish to link these mention spans against a \textit{biomedical knowledge graph} (\biokg), where medical knowledge is structured and systematised \cite{Kartchner2023ACE}.

While there are datasets available to train and benchmark clinical and biomedical EL models, 
 e.g., SNOMED-CT EL challenge for clinical notes~\cite{Davidson2025-kj}, BC5CDR for chemical–disease literature~\cite{b5cdr}, or COMETA for social media posts~\cite{basaldella-etal-2020-cometa}, existing datasets  suffer from three critical issues.
 \textbf{Semantic fragmentation:} Datasets are anchored to a single BioKG (e.g., SNOMED-CT) or include texts from a single domain (e.g., clinical notes). This creates information siloes leading to models that will not generalise beyond their biomedical vocabulary/domain.
\textbf{Explainability:} ``Black-box'' models increasingly face regulatory push-back in safety-critical domains \cite{huang2024explainable, ullah2024challenges}. There is a distinct lack of ground-truth data to train and evaluate interpretable clinical NLP models, especially for EL and similar tasks.
\textbf{Superficial evaluation}: The performance of EL models is typically measured using ``flat'' metrics like precision, recall, and F1-score. While useful insofar,
these metrics treat all errors as equal, e.g., incorrectly linking \textit{congestive heart failure} to \textit{myocardial infarction} (both types of heart disease) is penalized identically to linking it to \textit{influenza} (a completely unrelated viral disease). In other words, such metrics fail to capture the semantic nuance of prediction errors and do not distinguish models that make more plausible mistakes \cite{falis2021cophe, Amig2022EvaluatingEH, Plaud2024RevisitingHT}.

In this work, we introduce \dataset{}, a large-scale Entity Linking dataset that addresses all the above issues.
\dataset{}'s main features include:
\begin{itemize}
    \item \textbf{Integration}: We harmonise and integrate \textit{nine expert-annotated, curated datasets} covering clinical notes (ShARe/CLEF 2013,~\citealt{10.1007/978-3-642-40802-1_24}; SNOMED-CT EL Challenge,~\citealt{Davidson2025-kj}), biomedical literature (BC5CDR,~\citealt{b5cdr}; NCBI Disease,~\citealt{Dogan2014-cu}; MedMentions,~\citealt{mohan2019medmentionslargebiomedicalcorpus}), drug-label prose (TAC 2017 ADR,~\citealt{roberts2017overview}), and social media (CADEC,~\citealt{Karimi2015-we}; COMETA,~\citealt{basaldella-etal-2020-cometa}), totalling 500,000+ mentions and 45,000 unique concepts.
    \item \textbf{Vocabulary normalization}: We normalise all entities, grounded in different \biokgs{}, to a canonical UMLS CUI \citep[2025 AA;][]{Bodenreider2004TheUM}. We also map each code in one vocabulary to corresponding codes in all other covered vocabularies (up to 62 vocabularies in total).\footnote{We use the terms \textit{biomedical knowledge graph}, \textit{controlled clinical vocabulary}, and \textit{vocabulary} interchangeably.}
    \item \textbf{Hierarchical multi-vocabulary paths}: We annotate each concept with full hierarchical paths (i.e., from coarser to finer concepts) for the vocabularies that expose a usable API or where the full hierarchy is publicly available for download (11 biomedical vocabularies in total).
\end{itemize}

In Figure \ref{fig:overview}, we show the step-by-step process consisting of UMLS normalisation, vocabulary mapping, and hierarchical annotation generation. We show two running examples from the NCBI-Disease and CADEC datasets, respectively, clearly illustrating how our dataset addresses the semantic fragmentation and hierarchical annotation gaps. 

In Sections~\ref{sec:experiments} and~\ref{sec:results}, we introduce hierarchy-aware evaluation metrics (exact, ancestor-based, and hierarchy-based) and show initial experiments using \dataset{} on vocabulary-agnostic EL.

Finally, we release the codebase to reproduce \dataset{} under a permissive open-source licence at \url{https://github.com/mnishant2/MedPath}.

\section{Related work}
\label{sec:related}

\begin{table*}[t!]
\centering
\resizebox{\textwidth}{!}{%
\begin{tabular}{l c c c r l r r r c c c}
\toprule
\textbf{Dataset} / \textbf{Benchmark} & \textbf{Release} & \textbf{NER} & \textbf{EL} & \textbf{\#items} & \textbf{Unit} & \textbf{\#datasets} & \textbf{\#tasks} & \textbf{\#vocabs} & \textbf{Vocab I\&S} & \textbf{Path ann.} & \textbf{TUI ann.} \\
\midrule
\rowcolor{purple!10}
GERBIL platform              & 2015 & \textcolor{red}{\no}   & \textcolor{blue}{\yes}  & —         & —            & 32   & 1 (EL eval)          & 5+    & \textcolor{red}{\no}        & \textcolor{red}{\no}  & \textcolor{red}{\no} \\
MedMentions                  & 2019 & \textcolor{blue}{\yes}  & \textcolor{blue}{\yes}  & 352k & mentions     & 1    & 2 (NER, EL)          & 1     & \textcolor{red}{\no}        & \textcolor{red}{\no}  & \textcolor{red}{\no} \\
\rowcolor{purple!10}
BLUE                & 2019 & \textcolor{blue}{\yes}  & \textcolor{red}{\no}   & —         & —            & 10   & 5 (NER, RE, QA) & 0  & n/a        & \textcolor{red}{\no}  & \textcolor{red}{\no} \\

CrossNER                     & 2020 & \textcolor{blue}{\yes}  & \textcolor{red}{\no}   & 5{,}318   & paragraphs   & 1    & 1 (NER)              & 0     & n/a        & \textcolor{red}{\no}  & \textcolor{red}{\no} \\
\rowcolor{purple!10}
Few-NERD                     & 2021 & \textcolor{blue}{\yes}  & \textcolor{red}{\no}   & 491k & entities     & 1    & 1 (NER)              & 0     & n/a        & \textcolor{red}{\no}  & \textcolor{red}{\no} \\

BLURB                        & 2021 & \textcolor{blue}{\yes}  & \textcolor{red}{\no}   & —         & —            & 13   & 6 (LU tasks)         & 0     & n/a        & \textcolor{red}{\no}  & \textcolor{red}{\no} \\
\rowcolor{purple!10}
BigBio                       & 2022 & \textcolor{blue}{\yes}  & \textcolor{blue}{\yes} & $\sim$24M & examples    & 126+ & 13 categories       & 5+     & \textcolor{orange}{P} & \textcolor{red}{\no} & \textcolor{red}{\no} \\

BELB                         & 2023 & \textcolor{red}{\no}   & \textcolor{blue}{\yes}  & 347k & mentions     & 11   & 1 (EL)               & 7     & \textcolor{orange}{P} & \textcolor{red}{\no} & \textcolor{red}{\no} \\
\rowcolor{purple!10}
MedInst                      & 2024 & \textcolor{red}{\no}   & \textcolor{red}{\no}   & 7M & instructions & 133  & 133 (instr.)        & 8+     & n/a        & \textcolor{red}{\no} & \textcolor{red}{\no} \\

BRIDGE                       & 2025 & \textcolor{blue}{\yes}  & \textcolor{red}{\no}   & 1.4M & samples     & 87   & 8 (clin. NLP)        & 0     & n/a        & \textcolor{red}{\no} & \textcolor{red}{\no} \\
\midrule
\rowcolor{purple!10}
\textbf{\dataset{} (ours)}   & 2025 & \textcolor{blue}{\yes}  & \textcolor{blue}{\yes}  & $\sim$512K & mentions     & 9    & 3$^\dagger$ & 62$^\ddagger$ & \textcolor{blue}{\yes} & \textcolor{blue}{\yes} & \textcolor{blue}{\yes} \\
\bottomrule
\end{tabular}}
\caption{Comparison of multi-dataset / benchmark resources.  
Symbols: {\textcolor{blue}{\textcolor{blue}{\textcolor{blue}{\yes}}}} = yes; {\textcolor{red}{\no}} = no; {\textcolor{orange}{P}} = partial/limited support.  
\#items counts gold-annotated units; “Unit” clarifies their type.  
\textbf{Vocab I\&S}: cross-ontology vocabulary \emph{integration \& standardization}.  
\textbf{Path ann.}: ancestor / hierarchy paths provided.  
\textbf{TUI ann.}: UMLS Semantic Type identifiers attached. $^\dagger$: NER, EL and hierarchical EL. $^\ddagger$: flat codes, and 11 vocabularies with full hierarchy} 
\label{tab:dataset_comparison}
\end{table*}

NER and EL are among the most important tasks in clinical NLP.
NER helps us identify all mention spans of clinically relevant entities in some input text.
In EL, the goal is to link these mention spans against a specific structured biomedical knowledge graph, e.g. SNOMED-CT.

Biomedical EL systems have evolved from lexical matchers like MetaMap \cite{aronson2010overview} and TaggerOne \cite{Leaman2016TaggerOneJNA}, to embedding-based retrievers such as SapBERT \cite{liu2020self} and BioSyn \cite{sung2020biomedical}, and finally to generative architectures \citep{de2020autoregressive,xiao2023instructed, Yuan2022GenerativeBEA}. 

Recent work has emphasized the critical need for interoperable biomedical NLP systems that are robust to vocabulary fragmentation. For instance, \citet{neumann2019scispacy} and \citet{beltagy2019scibert} highlight the challenge of deploying models across datasets grounded in different \biokgs{} and domains. Similarly, \citet{wadden2019entity} and \citet{fries2022bigbio} underscore the brittleness of vocabulary-specific pipelines, which limit generalization across real-world settings. \citet{zu2024pathel} proposes a collective entity linking method based on relationship paths, \citet{MAG}, and \citet{doser} demonstrate a knowledge base agnostic entity linking system, while \citet{jannet2014eter} introduced a novel metric for evaluation of hierarchical NER. 



\subsection{Task-specific corpora for NER and EL}
Early biomedical NER efforts relied on single-vocabulary, single-domain corpora.
Some notable examples include i2b2/VA 2010 linked against SNOMED CT \cite{i2b2_2010}, 
BC5CDR chemical–disease abstracts \cite{b5cdr} linked against MeSH, ShARe/CLEF 2013 clinical notes \cite{shareclef2013} linked against UMLS, and MedMentions \cite{mohan2019medmentionslargebiomedicalcorpus} densely annotated PubMed abstracts against UMLS.

Later corpora widened the source spectrum, e.g. TAC2017  \cite{tac2017adr} that included drug-label prose (MedDRA)  and COMETA \cite{basaldella-etal-2020-cometa} that had social-media posts (SNOMED CT).  Individually, they cover a diverse range of data sources and formats, annotation guidelines, entity types, and native controlled clinical vocabularies. However, since they are anchored to a single vocabulary and a bespoke span guideline,  it is difficult to harmonize model implementation and/or benchmarking across them, leading to a lack of interoperability.
\subsection{Large-Scale Biomedical Benchmarks}
Recently, efforts have been made to consolidate individual datasets into larger task-based benchmark suites, which aim to homogenize fragmented datasets in terms of volume and diversity \cite{he2023medeval, Rouhizadeh2024ADFA}.
\textbf{BLURB} \cite{blurb} is a composite dataset that bundles 13 biomedical language understanding tasks (sentence similarity, NER, QA, etc.), but it does not target entity linking, provides no unified concept identifiers, and focuses almost exclusively on literature.   
\textbf{BigBio} \cite{fries2022bigbio} is a wrapper that brings together over 120 different biomedical datasets, including NER and NED based corpora, and streamlines them into a common Huggingface schema, but keeps original IDs and offers no cross-vocabulary mapping.  
\textbf{BELB} \cite{garda2023belb} is another collated large-scale dataset that is specifically concerned with Entity Linking. It includes 11 different EL datasets with 7 knowledge bases into one shared leaderboard with thorough benchmarking, still evaluating “flat” CUI accuracy within the native KB of each corpus.  
\textbf{MedInst} \cite{han2024medinst}, the newest, LLM-focused entrant among large BioMedical benchmarks, repurposes 130+ datasets for LLM instruction-tuning, again without normalising concept identifiers or exposing vocabulary structure.
\textbf{GERBIL} \cite{usbeck2015gerbil}, while not a dataset, deserves a mention when talking about clinical NER and EL. It is a web-based benchmarking framework that provides a web API for EL evaluation but ships no harmonised data or hierarchy metadata.\\ \\
\vspace{-0.1\baselineskip}
In Table \ref{tab:dataset_comparison}, we compare \dataset{} to well-known benchmarks across a breadth of capabilities. 
No existing benchmark simultaneously provides: 
(i) \emph{cross-vocabulary integration and standardization} (from UMLS CUIs to codes in up to 62 KBs) and  
(ii) \emph{explicit hierarchical paths in up to 11 vocabularies} for explainable model training and/or evaluation beyond flat metrics such as F1 score.

\subsection{Hierarchy-aware Entity Linking}

Entity linking and multi-label classification over \biokg{}-based label spaces require evaluation metrics that provide partial credit for semantically similar predictions rather than treating near-miss predictions as complete failures. The foundational work by \citet{kosmopoulos2015evaluation} provides a comprehensive, unified framework for hierarchical evaluation, introducing LCA-based (Lowest Common Ancestor) metrics that construct minimal graphs connecting predicted and true labels via their LCAs, thereby avoiding over-penalization at deeper hierarchy levels. Earlier approaches \cite{kiritchenko2005functional, kiritchenko2006learning} augment predictions with all ancestor classes, while the CoPHE metric \cite{falis2021cophe} preserves count information during ancestor propagation, enabling detection of over- and under-prediction within label families. The H-loss framework \cite{cesa2006incremental} charges loss only for the first classification mistake along prediction paths, capturing the intuition that coarse-grained errors subsume fine-grained mistakes, though limited to tree-structured hierarchies.

Despite the rich hierarchical structures in biomedical terminologies such as UMLS (127 semantic types) and SNOMED-CT (364K concepts in DAG structure), hierarchical evaluation remains notably absent from entity linking assessment. \citet{kartchner2023comprehensive} demonstrates that major biomedical entity linking datasets rely exclusively on flat metrics, basic accuracy, relaxed matching, and strict matching, with no use of hierarchical partial credit, despite vocabularies providing is-a, part-of, and definitional relationships that could inform evaluation. \citet{pesquita2009semantic} survey content-based semantic similarity measures extensively used in Gene Ontology applications, yet these remain underutilized in entity linking evaluation. \citet{kosmopoulos2015evaluation} report that metric selection can fundamentally alter system rankings and reveal distinct error patterns such as over-/under-specialization and sibling confusion, which flat metrics treat identically but have vastly different downstream consequences in clinical decision support applications. \dataset{} addresses this evaluation gap by providing hierarchical path annotations for 11 vocabularies, in addition to flat single-concept identifiers, explicitly capturing the ancestor lineage from root to leaf concepts. This design facilitates granular analysis of hierarchical evaluation metrics and informed modeling choices. In our initial experiments, we employ basic hierarchical metrics, including ancestor and descendant accuracy, to illustrate the utility of \dataset{} for path-based evaluation.

\section{Dataset Construction}
\label{sec:methods}

\begin{table*}[t!]            
\centering
\scriptsize
\begin{tabularx}{\textwidth}{
  >{\bfseries}p{3.1cm}  
  c                     
  Y                     
  Y                     
  Y                     
  Y}                    
\toprule
Dataset & \textbf{Year} & \textbf{Domain / Source (docs)} & \textbf{Entity types} & \textbf{Ontology} & \textbf{Licence}\\
\midrule
\rowcolor{purple!10}
\makecell[l]{SNOMED CT EL Challenge\\\cite{Davidson2025-kj}} & 2023 & MIMIC-IV ICU discharge notes (300) & Disorder, Procedure, Drug, Device & SNOMED CT & PhysioNet-R-A \\ [2pt]
\makecell[l]{ShARe/CLEF 2013 \\\cite{10.1007/978-3-642-40802-1_24}} & 2013 & Hospital discharge notes (199) & Disorder, Procedure, Medication, Device & UMLS & PhysioNet DUA \\[2pt]
\rowcolor{purple!10}
\makecell[l]{Mantra GSC (English) \\ \cite{10.1093/jamia/ocv037}} & 2015 & Patents, drug labels, abstracts (1 050) & 16 UMLS semantic groups & UMLS subset & CC-BY-SA 4.0 \\[2pt]
\makecell[l]{BC5CDR \\\cite{b5cdr}} & 2016 & PubMed abstracts (1 500) & Chemical, Disease & MeSH & CC-BY 3.0 \\[2pt]
\rowcolor{purple!10}
\makecell[l]{NCBI Disease \\\cite{Dogan2014-cu}} & 2014 & PubMed abstracts (793) & Disease & MeSH, OMIM & CC-BY 4.0 \\[2pt]
\makecell[l]{MedMentions \\\cite{mohan2019medmentionslargebiomedicalcorpus}} & 2019 & PubMed abstracts (4 392) & Any UMLS concept & UMLS & CC-BY 4.0 \\[2pt]
\rowcolor{purple!10}
\makecell[l]{TAC ADR 2017 \\\cite{roberts2017overview}} & 2017 & FDA Structured-Product Labels (200) & ADR, Drug & MedDRA, RxNorm & Public domain \\[2pt]
\makecell[l]{CADEC \\\cite{Karimi2015-we}} & 2015 & Patient forum posts (1 250) & ADE, Drug & MedDRA, SNOMED CT & Ask-A-Patient T\&C \\[2pt]
\rowcolor{purple!10}
\makecell[l]{COMETA \\\cite{basaldella-etal-2020-cometa}} & 2020 & Reddit + Twitter posts (20 000) & Symptom & SNOMED CT & CC-BY-NC \\[2pt]
\bottomrule
\end{tabularx}
\caption{\textbf{\dataset{} core datasets}.  Detailed statistics (mentions, CUIs, path depth) appear in Table~\ref{tab:dataset-overview}.}
\label{tab:harmony_core}
\end{table*}

\subsection{Curation rationale and source datasets}
We first conducted a comprehensive survey of biomedical and clinical corpora available from institutional, shared task, and open-source repositories. Our primary selection criterion was the presence of high-quality, expert-validated ground-truth annotations suitable for EL. To ensure the final resource would be a challenging testbed for model generalization, we also prioritized datasets that collectively offered maximum diversity in textual domains and semantic types.

The nine corpora selected through this curation process are detailed in Table \ref{tab:harmony_core}. While not an exhaustive representation of the ever-evolving biomedical field, this collection constitutes a large-scale and domain-diverse resource. It spans a wide spectrum, from formal scientific literature and clinical notes to product labels and informal social media content. Overall, the unified corpus comprises over 5 million tokens, more than 500,000 entity mentions, and 45,000 unique concepts, all drawn from source datasets with permissive licenses for research use. 

\subsection{Annotation}
\label{sec:annotation}

\dataset{} was created with a four-stage automated pipeline. This process integrates the individual datasets with fragmented annotations into a single, cohesive, and multi-vocabulary benchmark.

\paragraph{Stage 1: Unification and Standardization} The nine source corpora are published in disparate formats, including BRAT, PubTator, XML, and TSV. Our first step was to develop dataset-specific parsers to ingest these formats and convert them into a standardized JSON schema. This process also involved light text cleaning to remove artifacts (e.g., de-identification remnants, stray characters) while preserving the original annotations.

\paragraph{Stage 2: Canonicalization via UMLS Mapping} To resolve semantic fragmentation, we normalized all concept IDs from their native \biokg{} to the latest Unified Medical Language System (UMLS 2025AA) release.
For each mention, we first attempt to map from its vocabulary native ID directly to a UMLS Concept Unique Identifier (CUI) using a dictionary created from the UMLS database. If this fails, we fallback first to an exact match and then to a semantic containment heuristics.\footnote{Exact match: we string match between the mention text against each concept name in the UMLS term dictionary. If no exact match is found, we check for bidirectional substring containment (i.e., \texttt{X in Y} or \texttt{Y in X}). We use all concept names and synonyms available for each CUI, and choose the closest match based on token overlap and length similarity.}
This fallback strategy could introduce noise in the annotations, but the information loss from discarding these examples is a decrease in 2.5\% in the number of unique mentions (2.13\% from exact match, 0.37\% from semantic containment) and 1.15\% in the number of unique concepts (0.81\% exact match, 0.35\% semantic containment).
In absolute numbers, we have 513,218 mentions / 44,259 unique concepts (CUIs) including examples mapped via exact match and semantic containment, and 500,384 mentions / 43,396 unique concepts after excluding these examples.

\paragraph{Stage 3: Multi-level Semantic Enrichment} With a canonical CUI for each mention, we added two further layers of semantic information. First, we extracted the corresponding Semantic Type (TUI) for each unique CUI, providing a high-level categorization for every entity that was used in our initial experiments (see Section~\ref{sec:experiments}). Second, to enable interoperable, vocabulary-agnostic research, we mapped each CUI to its parallel concept identifiers in other major biomedical vocabularies, leveraging the atom-level information within UMLS.

\paragraph{Stage 4: Hierarchical Path Extraction} The final and arguably most important stage of our pipeline was the extraction of full hierarchical paths, whereby we provide a data structure encoding rich hierarchical information to enable novel applications. This was a multi-step process. First, we identified the top 25 most frequently represented vocabularies in the datasets we use (Table~\ref{tab:dataset-overview}) and determined which of them possessed both \textit{a formal hierarchical structure} and \textit{an accessible native taxonomy} (via public API or downloadable files). This process yielded 11 target vocabularies for path extraction. All the vocabularies we surveyed are listed in the Appendix \ref{sec:vocabularies_appendix}.

\vspace{-0.1\baselineskip}
Next, we developed a custom path extraction method for each of these 11 vocabularies. This involved creating bespoke extractor modules that respect the unique structure of each vocabulary (e.g., interpreting different relationship types such as \texttt{is-a} relations in SNOMED CT or tree numbers in MeSH). For each concept, the extractor iteratively traverses the hierarchy from the entity to its more general/parent terms until either a root node is reached or no parent node can be found. The process was designed to be exhaustive, capturing and storing all possible paths for concepts that exist in multiple inheritance structures, i.e., when a vocabulary allows for more than one parent per node. Wherever presented with a choice, the latest version of the vocabulary was selected during this step. To ensure scalability and robustness, the implementation included several technical optimizations, such as result caching, filtering of inactive or obsolete codes, and robust API callback handling. This final stage produced a total of 573,786 distinct hierarchical paths for the 44,259 unique concepts.
\subsection{Data Schema} 
An illustrative example of our multi-layered annotation schema is presented in Figure \ref{fig:overview}, with the full JSON specifications detailed in Appendix \ref{sec:dataset_schema}.

\subsection{Data Quality Validation}
\label{sec:data_quality_validation}
We had a strict requirement to select only datasets with clinical expert oversight involved in the data curation. Moreover, we implemented multiple layers of checks and validations at each stage of the automated workflow to ensure data quality. The only stage where automatic heuristics may introduce noise is in mapping from source vocabulary codes to UMLS CUIs, and only when no mapping of native ID-to-CUI exists.
We explain how we address this in Section~\ref{sec:annotation}, Stage 2. In short: possibly noisy examples mapped via exact match and semantic containment are clearly labelled in \dataset{}, meaning that users can either filter them out and have a noise-free dataset, or use them in case their use-case allows. 

\subsection{Availability and Update Strategy}
In the interest of reproducibility, we release our complete annotation pipeline, data processing scripts, and evaluation code under a permissive open-source license. However, several of the constituent datasets and source vocabularies that form \dataset{} are protected by their own licenses or data usage agreements (DUAs) and cannot be redistributed directly. In such cases we provide detailed instructions and scripts that allow researchers who have obtained the necessary permissions from the original data providers to apply our pipeline and fully reconstruct \dataset{}. Keeping in mind that controlled clinical vocabularies are living resources which undergo frequent updates, we implemented the data preprocessing and annotation in a way that MedPath’s scripts are easily compatible with any version of the various resources used (e.g., UMLS, SNOMED, MedDRA, etc). For example, we currently use \textit{UMLS 2025 AA}, \textit{SNOMED CT May 2025}, and \textit{MedDRA 27.1}. However, one can easily adjust these versions by changing a single parameter/argument in the code to generate MedPath with future versions of these vocabularies. 

\section{Analysis}
\label{sec:analysis}
\definecolor{headercolor}{RGB}{255, 255, 255}
\definecolor{rowcolor1}{RGB}{236, 240, 241}
\definecolor{rowcolor2}{RGB}{255, 255, 255}
\definecolor{SciAbs}{HTML}{66C2A5}
\definecolor{Clin}{HTML}{FC8D62}
\definecolor{SocMed}{HTML}{8DA0CB}
\definecolor{Patent}{HTML}{E78AC3}
\definecolor{Mixed}{HTML}{A6D854}

\begin{table}[t!]
\centering
\caption{Biomedical entity linking datasets.  Domain codes: 
{\color{SciAbs}\textbf{SA}}\,=Scientific Abstracts;\,
{\color{Clin}\textbf{CN}}\,=Clinical Notes;\,
{\color{SocMed}\textbf{SM}}\,=Social Media;\,
{\color{Patent}\textbf{DP}}\,=Drug Patents;\,
{\color{Mixed}\textbf{MX}}\,=Mixed.}
\label{tab:dataset-overview}
\resizebox{0.48\textwidth}{!}{%
\begin{tabular}{lrrrrr}
\toprule
\rowcolor{headercolor}
{\textbf{Dataset}} & 
{\textbf{Docs}} & 
{\textbf{Mentions}} & 
{\textbf{CUIs}} & 
{\textbf{TUIs}} &  
{\textbf{Domain}} \\
\midrule
\rowcolor{purple!10}
MedMentions  & 4392   & 352496 & 34631 & 126 & {\color{SciAbs}\textbf{SA}}\\
\rowcolor{rowcolor2}
MIMIC-IV-EL  & 204    & 51574  & 5258  & 52 & {\color{Clin}\textbf{CN}}\\
\rowcolor{purple!10}
TAC 2017 ADR & 200    & 32585  & 3098  & 94 & {\color{Patent}\textbf{DP}}\\
\rowcolor{rowcolor2}
BC5CDR       & 1500   & 29076  & 2487  & 69 & {\color{SciAbs}\textbf{SA}}\\
\rowcolor{purple!10}
COMETA       & 20015  & 20015  & 3864  & 82 & {\color{SocMed}\textbf{SM}}\\
\rowcolor{rowcolor2}
CADEC        & 1186   & 9842   & 1256  & 68 & {\color{SocMed}\textbf{SM}}\\
\rowcolor{purple!10}
ShaRe/CLEF   & 291    & 8676   & 1372  & 34 & {\color{Clin}\textbf{CN}}\\
\rowcolor{rowcolor2}
NCBI-Disease & 792    & 7026   & 741   & 35 & {\color{SciAbs}\textbf{SA}}\\
\rowcolor{purple!10}
Mantra-GSC   & 526    & 1928   & 1276  & 92 & {\color{Mixed}\textbf{MX}}\\
\midrule
\rowcolor{headercolor}
{\textbf{Overall}} & 
{\textbf{29,106}} & 
{\textbf{513,218}} & 
{\textbf{44,259}} & 
{\textbf{126}} & 
{\textbf{4}} \\
\bottomrule
\end{tabular}}
\label{tab:dataset_composition}
\end{table}

\begin{figure}[t!]
  \centering
  \includegraphics[width=\linewidth]{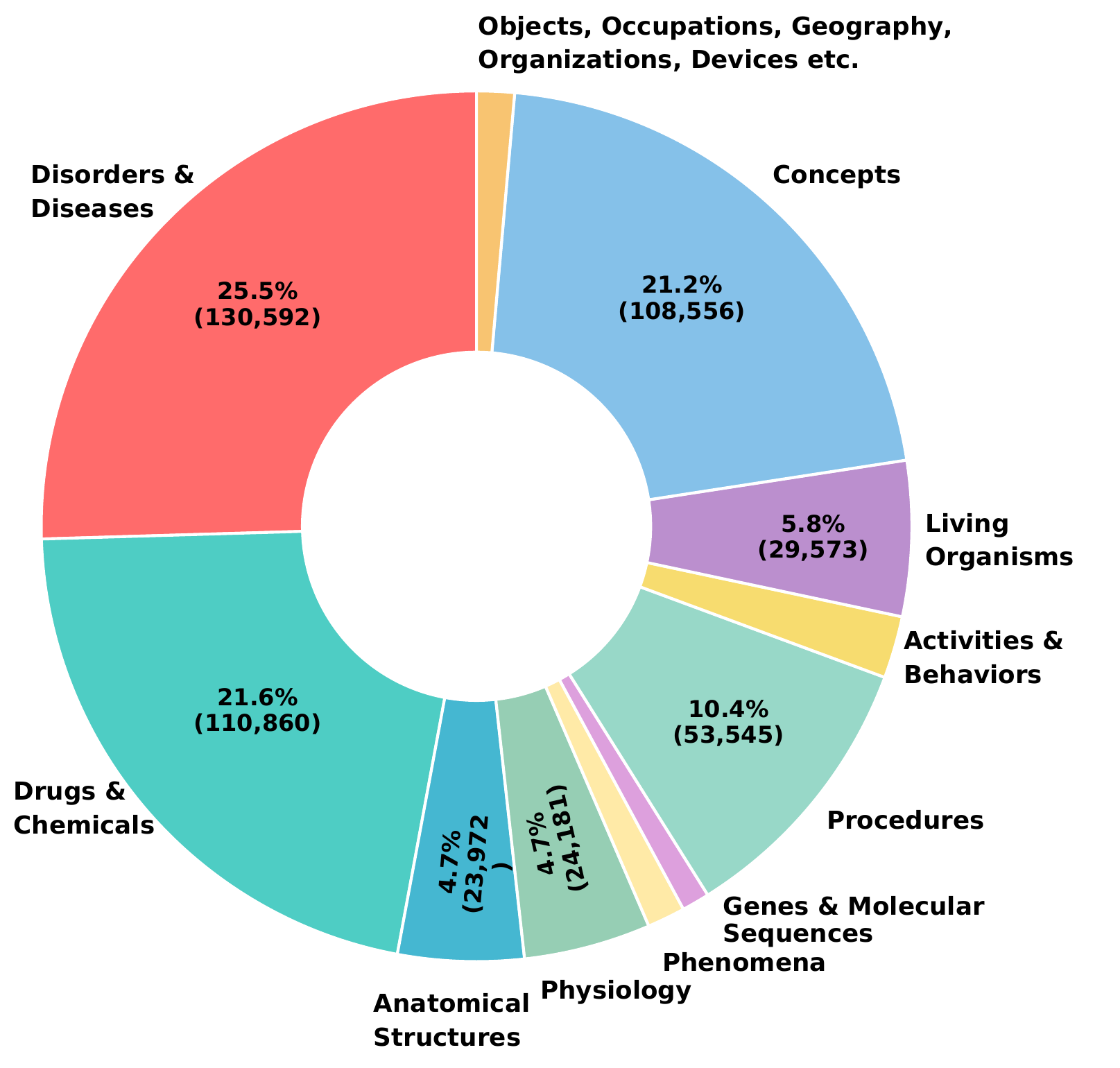}
  \caption{Semantic type distribution in \dataset{}.}
  \label{fig:sem_type}
\end{figure}

To characterize the properties of \dataset{}, we conducted a detailed statistical analysis. The following sections quantify the dataset's scale, its conceptual breadth, and the richness of its semantic and hierarchical annotations. More detailed analysis can be found in Appendix \ref{sec:Dataset_appendix}.
\subsection{Size and Genre Balance}
The final harmonized corpus comprises over 5 million tokens and 513k expert-annotated mentions (Table \ref{tab:dataset_composition}). While MedMentions easily dominates the mentions count, the fact that it is itself a great mix of biomedical text is crucial for the diversity of our dataset.
Social media posts add 20\% of documents but only 6\% of mentions, illustrating their short-form nature and motivating cross-length generalisation.
For a more granular analysis, we categorized the source corpora into four primary domains based on their source as detailed in Table \ref{tab:harmony_core}.
Scientific literature (MedMentions, BC5CDR, NCBI-Disease) constitutes the largest portion, a reflection of its relative accessibility and textual density.
The other three domains---clinical notes (MIMIC-IV-EL, ShARe/CLEF), social media (COMETA, CADEC), and drug labels and patents (ADR)---are well-balanced and contribute significant domain-specific richness.
The Mantra-GSC corpus, which contains text from Medline abstracts, drug labels, and patent claims, was classified as a mixed-domain dataset.

\subsection{Concept Breadth and Semantic Diversity}
Across datasets we observe a rich mix of concepts.
Whereas all datasets combined have a mapped UMLS concept count of $\sim$$54,000$, the unique mapped CUIs are $44,259$. This indicates a concept overlap of only about 20\% across the datasets, underscoring the value of harmonization for creating a comprehensive benchmark that moves beyond the semantic scope of any single source. 
The semantic diversity of the corpus is equally broad. The annotated mentions span 126 of the 127 possible high-level UMLS Semantic Types (STYs). The most prominent semantic groups are \textit{Disorders and Diseases} (25.5\%), \textit{Drugs and Chemicals} (21.6\%), and \textit{concepts} (21.2\%). See Figure \ref{fig:sem_type} for details.

\subsection{Vocabulary Coverage and Hierarchy Insights}
\begin{figure}[t!]
  \centering
  \includegraphics[width=\linewidth, height=0.25\textheight]{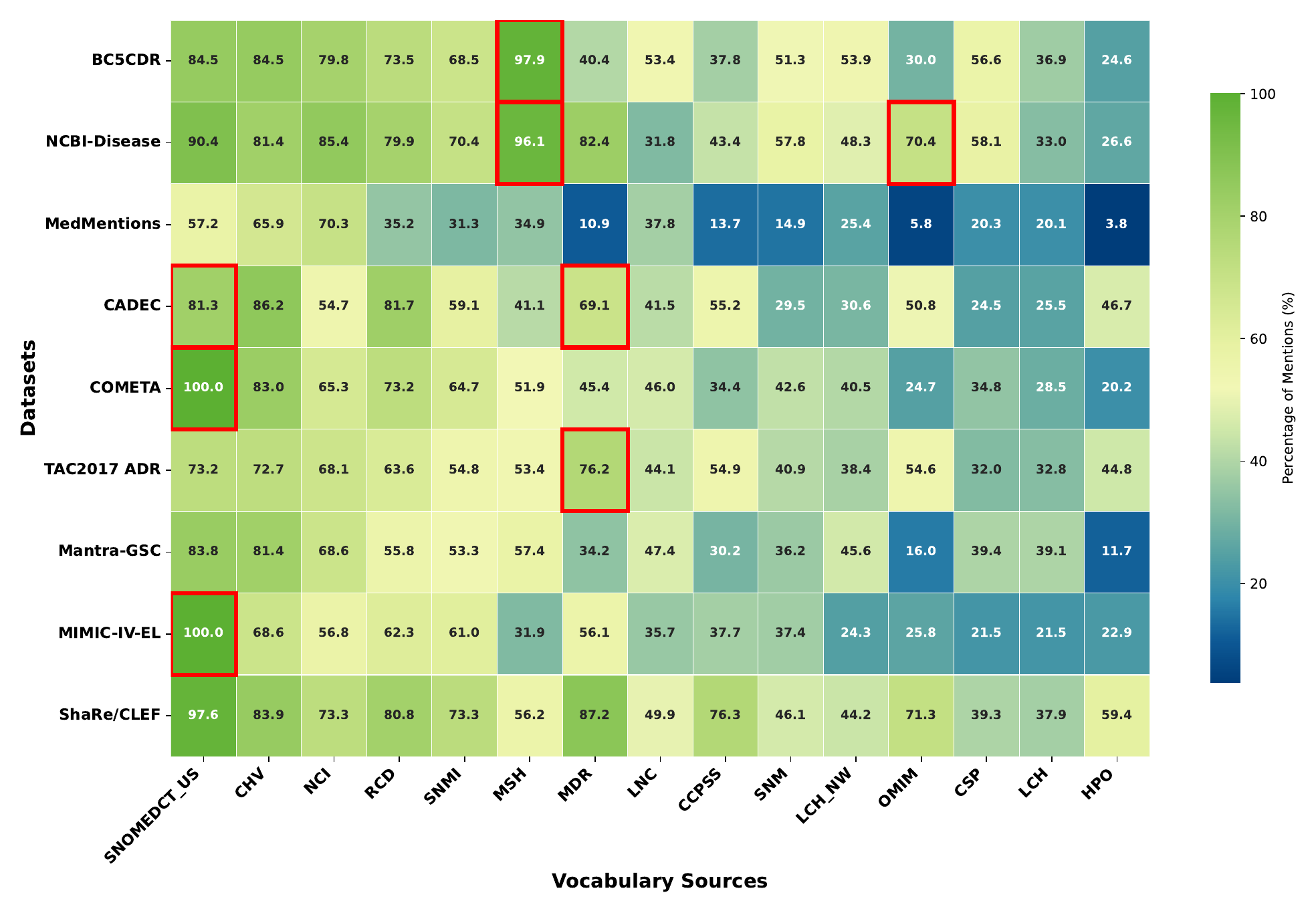}
  \caption{Vocabulary overlap heat map. Datasets' annotations using UMLS are not shown.}
  \label{fig:heatmap}
\end{figure}

Figure \ref{fig:heatmap} illustrates the extensive cross-vocabulary coverage of the resource, displaying the distribution of mentions from each source dataset across the 15 most frequent vocabularies. This visualization highlights the degree of interoperability achieved through our normalization pipeline. A key finding is the centrality of SNOMED-CT; seven of the nine datasets map over 80\% of their mentions to SNOMED-CT concepts, even those with different native knowledge bases, demonstrating its comprehensive integration within UMLS.

\begin{figure}[t!]
  \centering
  \includegraphics[width=\linewidth]{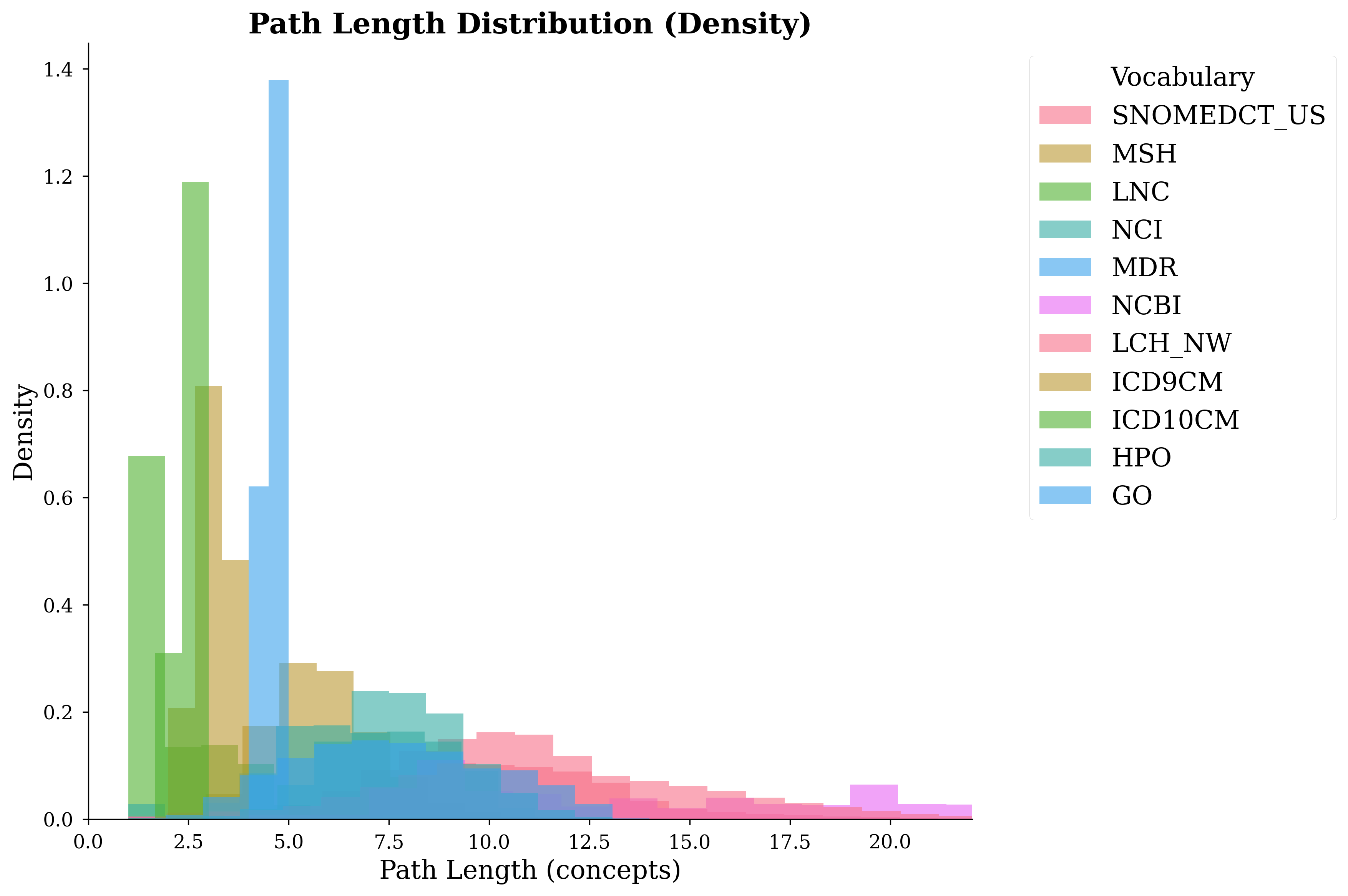}
  \caption{Histogram of lengths of entity hierarchical paths across different vocabularies.}
  \label{fig:path_length}
\end{figure}

Beyond simple coverage, we analyzed the structural properties of the 11 vocabularies for which we extracted full hierarchical paths. SNOMED-CT is the most information dense and structurally complex; 84\% of its mapped concepts feature multiple inheritance paths, with an average of over 20 distinct paths per concept. The distribution of path depths, shown in Figure \ref{fig:path_length}, reveals significant diversity across ontologies. SNOMED-CT exhibits a wide spread of path lengths, NCBI contains the deepest hierarchies on average, while others like ICD-9, ICD-10, and MedDRA have more concentrated path lengths of 3--5 levels, consistent with their defined structures. This variety in granularity, from complex directed acyclic graphs, like SNOMED CT, to simpler tree structures, confirms that \dataset{} is well-suited for developing and evaluating coarse-to-fine, hierarchy-aware models.

\section{Preliminary Experiments}
\label{sec:experiments}
We now provide initial experiments showcasing performance gains obtained using \dataset{} compared to training models on individual datasets and on datasets from a single domain.
While \dataset{} can also be used for NER, our primary focus is on biomedical EL. We thus present EL experiments in this section and, for completeness, report on preliminary NER experiments in Appendix \ref{sec:baseline_appendix}.





\subsection{Biomedical Entity Linking}
We implement and benchmark a two-stage EL model adapted from the X-MEN library \cite{borchert2025xmen}.\footnote{\url{https://github.com/hpi-dhc/xmen/tree/main}}

\paragraph{Retrieval}
We adopt two lightweight, dictionary-based retrieval methods implemented using: (i) TF-IDF-vectorizer operating over character 3-grams based retrieval, and (ii) embedding-based retrieval with SapBERT \citep{sapbert}. Both methods index a unified dictionary built from UMLS CUIs and their associated \textit{names}, \textit{synonyms}, and \textit{lexical variants}. We include UMLS CUIs linked to any example from any source dataset in \dataset{}. Each test mention is treated as a query to retrieve the top-$k$ most similar CUIs from this index.

\paragraph{Reranking}
The retrieved candidates from both TF-IDF and SapBERT are then passed to a cross-encoder model that performs reranking to identify the most relevant entity. Cross-encoders were trained on the top-$32$ generated candidates plus the gold entity (in case the gold entity is not in the top-$32$).
We use a categorical cross-entropy loss function with regularization to optimize for ranking performance.
The model can be initialized from various pretrained BERT encoders; we used \texttt{cambridgeltl/SapBERT-from-PubMedBERT-fulltext}. The model is trained to maximize top-$1$ accuracy, with the checkpoint that achieves the highest validation accuracy being selected for final inference.

\paragraph{Evaluation}
We use a test set consisting of all unique mentions with ground-truth CUIs across datasets. We report \textbf{accuracy@$k$} ($k = 1, 5, 32$) and \textbf{mean reciprocal rank} (MRR).
To show the importance of semantically aware metrics for entity linking, we compute hierarchically-aware metrics that help assess both coarse and fine-grained performance of the models. For each mention whose gold CUI maps to one of 11 vocabularies with extracted hierarchies, which approximately covered all the mentions (98.7\%), we evaluate whether any of the top-$k$ predicted CUIs are (i) ancestors (\textbf{Ancestor@k}), (ii) descendants (\textbf{Descendant@k}), or (iii) part of a hierarchy in any way (\textbf{Hierarchy@k}) within the same \biokg{}, which could mean entities having common ancestors, or any hierarchy overlap with the test gold CUI, skipping top-3 levels from the root node of the vocabulary so we don't consider too general hierarchy match. 

\paragraph{Vocabulary-Agnostic Entity Linking}

In our first experiment, we benchmark TF-IDF and SapBERT-based retrievers across the full test set using only surface form matching against the UMLS-derived CUI dictionary. This vocabulary-agnostic retrieval simulates realistic scenarios where the mention surface form may originate from different vocabularies or domains.

\begin{table}[t!]
\centering
\caption{Overall Entity linking performance. “CG” = candidate generator, “+RR” = with reranker. 
Per row: best \underline{score in CG} underlined; best \textbf{score in CG+RR} bolded.}

\setlength{\tabcolsep}{3pt}
\renewcommand{\arraystretch}{1.05}
\footnotesize
\resizebox{\columnwidth}{!}{%
\begin{tabular}{lcccc}
\toprule
& \multicolumn{2}{c}{\textbf{CG}} & \multicolumn{2}{c}{\textbf{CG+RR}} \\
\cmidrule(lr){2-3}\cmidrule(lr){4-5}
\textbf{Metric} & \textbf{TF-IDF} & \textbf{SapBERT} & \textbf{TF-IDF} & \textbf{SapBERT} \\
\midrule
\multicolumn{5}{l}{\textbf{Standard Metrics}} \\
Acc@1   & \underline{51.46\%} & 48.12\% & \textbf{80.84\%} & 79.02\% \\
Acc@5   & 64.85\% & \underline{65.44\%} & 91.22\% & \textbf{92.60\%} \\
Acc@32  & 72.01\% & \underline{73.68\%} & 96.36\% & \textbf{98.76\%} \\
MRR@32  & \underline{0.5756} & 0.5594 & 0.857 & \textbf{0.861} \\
\midrule
\multicolumn{5}{l}{\textbf{Hierarchical Metrics}} \\
Hierarchy@1   & \underline{68.60\%} & 61.39\% & 85.40\% & \textbf{86.24\%} \\
Hierarchy@5   & \underline{82.56\%} & 80.66\% & 95.31\% & \textbf{96.30\%} \\
Ancestor@1    & \underline{20.73\%} & 18.74\% & \textbf{24.80\%} & 24.68\% \\
Ancestor@5    & \underline{27.58\%} & 25.16\% & 32.38\% & \textbf{34.02\%} \\
Descendant@1  & \underline{20.10\%} & 18.46\% & 23.45\% & \textbf{23.74\%} \\
Descendant@5  & \underline{29.42\%} & 25.39\% & \textbf{32.98\%} & 32.75\% \\
\bottomrule
\end{tabular}
}
\label{tab:el_overall}
\end{table}

\paragraph{Ablations}
In our ablation, we systematically compare training strategies for EL and NER reranking under three regimes: \textbf{in-dataset}, whereby train/test come from disjoint splits from a same dataset; \textbf{in-domain}, whereby we train on all but one dataset within a domain and test on that held-out dataset; and \textbf{overall} whereby we train on the union of all datasets across all domains,
i.e. \dataset{}.
The unified UMLS mapping enables consistent label semantics across datasets, letting us 1) pool supervision in the overall setting, 2) measure generalization across datasets within a domain, and 3) compute comparable per-type macro summaries.

\section{Preliminary Results and Discussion}
\label{sec:results}
\subsection{EL Experiment Results}
Table \ref{tab:el_overall} shows the performance of the two EL candidate generation methods using both standard and hierarchical metrics. Overall, TF-IDF outperforms SapBERT in accuracy (Acc@1 = 51.5\% vs. 48.1\%) and MRR. This suggests that lexical overlap remains a strong signal in biomedical entity linking, and high coverage of the dictionary built for linking.
SapBERT surpasses TF-IDF at k = 5 and k = 32 (Acc@5 = 65.4\%, Acc@32 = 73.7\%), indicating its strength in retrieving semantically similar or morphologically varied candidates not captured by character $n$-grams.
Adding a reranker on top drastically improves all metrics for both candidate generators, with the SapBERT generator plus the SapBERT reranker outperforming TF-IDF plus SapBERT reranker on most metrics.

Hierarchy-aware evaluation shows that our dataset enables a much richer analysis than exact CUI accuracy. While TF-IDF attains $\text{Acc@1} = 51\%$, its Hierarchy@1 jumps to 68.6\%, indicating that an additional $\sim$17\% of mentions retrieve a concept that is semantically related (i.e., a sibling, cousin, or ancestor). SapBERT exhibits a similar 13\% gain (48$\rightarrow$61\%). Roughly 20\% of errors are over-general (ancestor) and 20\% are over-specific (descendant), highlighting granularity ambiguity rather than synonym mismatch.

\begin{figure}[t!]
  \centering
  \includegraphics[width=\linewidth]{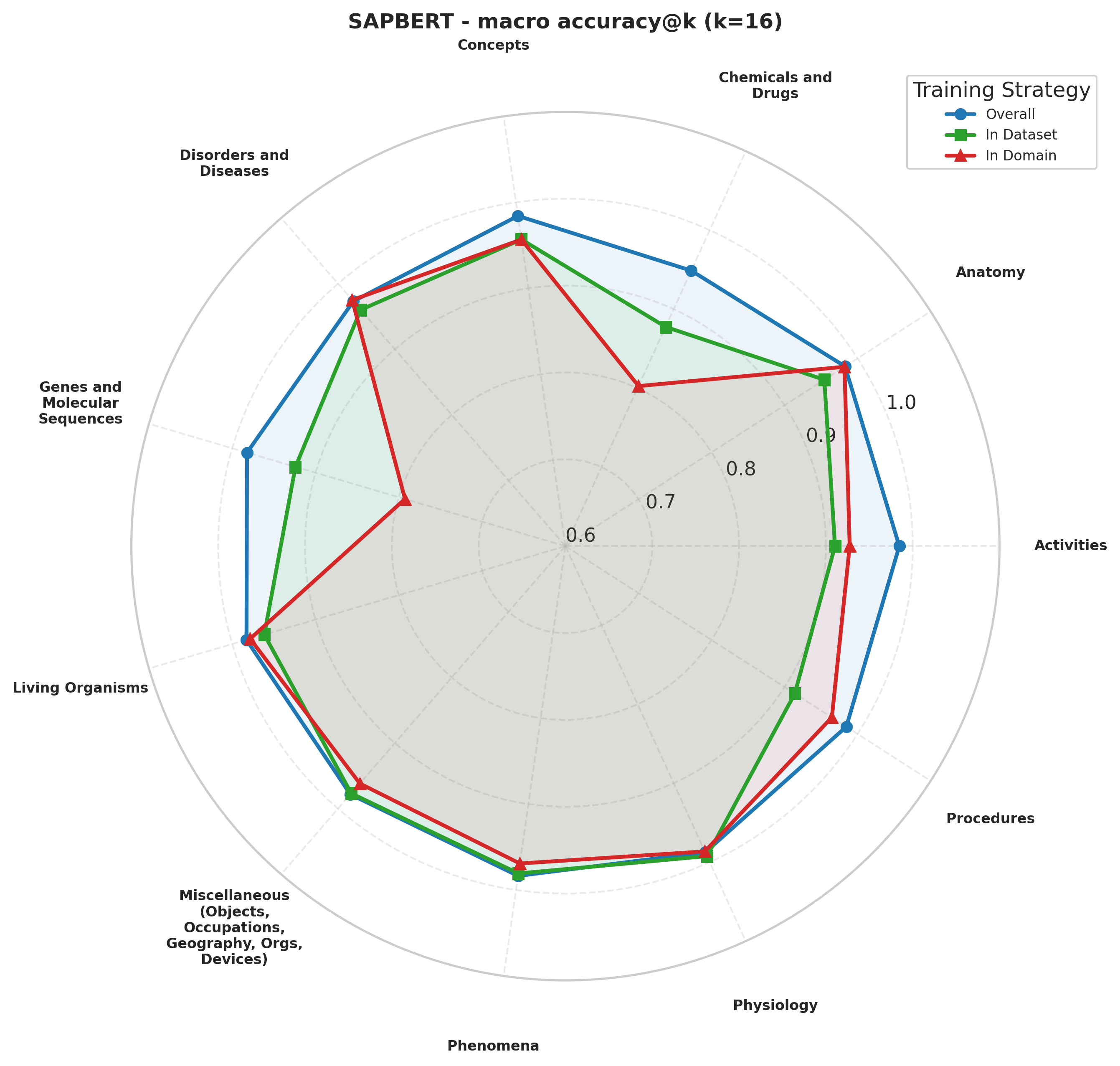}
  \caption{Figure showing EL performance in the three data settings}
  \label{fig:ablation}
\end{figure}

In Figure \ref{fig:ablation}, we see the reranker models performance when trained on a single dataset (in-dataset), on a single domain (in-domain), or on all datasets in \dataset{} (overall), in terms of macro-averaged acc@16 per semantic class. We observe that the model trained on \dataset{} consistently outperforms the other two settings, and sometimes by a large margin, thus validating the utility and information gain from collation and canonicalization in \dataset{}.

\section{Conclusions and Future Work}
\label{sec:conclusions}
In this work, we introduced \dataset{}, a large-scale resource for training and evaluating biomedical EL models that addresses three main limitations: semantic fragmentation, lack of explainability, and use of semantically-blind evaluation metrics. We integrate and harmonise nine diverse and expert-curated datasets across 4 domains with $513$k mentions and $45,000$ unique entities. We normalize all entity mentions to an up-to-date, canonical UMLS backbone, which means \dataset{} directly tackles the problem of data siloes.

\dataset{} includes mappings across up to 62 controlled clinical vocabularies and $\sim575$k hierarchical path annotations in 11 prominent clinical and biomedical knowledge graphs.
It enables the training and evaluation of inherently explainable NER and EL models, 
and
facilitates the development of truly diverse systems in terms of \biokg{}, a vital step towards achieving the interoperability required for real-world clinical deployment compatible with state-of-the-art generative AI methods. 
We release \dataset{} publicly at \url{https://github.com/mnishant2/MedPath} and hope to accelerate the development of biomedical NER and EL models that are more robust, trustworthy, and semantically aware.
\paragraph{Future work}
We believe \dataset{} opens research avenues in many directions.
\noindent
1) We can go \textit{beyond post-hoc explanation techniques and build inherently explainable models}.
We envision using \dataset{}'s hierarchical paths for training generative models that predict not only an entity's ID, but a mention's entire hierarchical path as a means to shed light on the model prediction process. 
2) \dataset{}'s vocabulary mapping across \biokgs{} allows for the construction of unified models that are \textit{fluent in several medical vocabularies}. Future work may investigate multi-task learning setups where a single
model is trained to make predictions across all
11 vocabularies. An EL system like this would be able to map a mention to its equivalent concepts in SNOMED-CT, MeSH, and ICD-10 all at once and would be a big step forward for model interoperability.
3) The hierarchical path annotations across 11 controlled clinical vocabularies provide a test-bed for the community to design and validate more sophisticated hierarchical evaluation metrics that can measure errors that encode domain-specific semantics within and across \biokgs{}.
4) Furthermore, hierarchical paths allow for fine-grained error analysis including answering questions like \textit{`Do models more frequently confuse sibling concepts more than distant ones?'} or \textit{`At what depth of the hierarchy do models begin to fail?'}
5) Additional applications of \dataset{} could include knowledge graph generation, and pre-training and fine-tuning LLMs so that LLMs are more factually grounded in established medical knowledge. 

\section*{Limitations}
While \dataset{} represents a significant step towards more diverse resources for biomedical NLP, we highlight a few limitations that users and researchers should be aware of. 

\paragraph{Diversity}
Although we merged nine corpora to achieve broad domain diversity, this collection is not exhaustive and does not represent the universe of biomedical and clinical text. Models trained on \dataset{} may not generalize well to text from under-represent sources, e.g., clinical notes from other Electronic Health Record (EHR) systems or from specialized sub-disciplines.
\dataset{} also only covers English datasets and does not address the needs of multilingual research in this domain. 

\paragraph{Annotation issues}
\dataset{}'s scale necessitated a largely automated pipeline for path extraction and entity normalization. While we employed state-of-the-art tools and devised a stringent methodology, with validation and statistical analysis, we did not perform expert, clinical validation of the new layers of annotation and mappings added.
Moreover, there may be errors inherited from the original datasets, e.g., incorrect entity links for highly ambiguous mentions, missing nested mentions, missing or misaligned mappings, incorrect/outdated codes.
Manual verification of mentions was not feasible. While we make the script with the annotation pipeline available to ensure transparency, users must be aware that the annotations reflect these limitations.

\paragraph{Versioning and updates}
Another potential area of concern and a well-known challenge is the ever-evolving nature of medical knowledge bases. Our annotations---UMLS CUIs, semantic types, and especially hierarchical paths---are tied to particular versions of the underlying ontologies, e.g., UMLS 2025 AA, MedDRA 27.0, SNOMED CT US March 2025.
Biomedical knowledge is, however, not static; these terminologies are continually updated, with concepts being added, deprecated, or redefined.
To this end, \dataset{} should be considered a high-fidelity snapshot at a particular point in time. As time passes, some paths will become outdated, and new concepts will not be represented, which may affect the resource's utility in the long term without periodic updates.
Although \dataset{}'s codebase makes it very easy to use a future version of a \biokg{} already in our pipeline, changes that break backward compatibility can still be an issue.
Moreover, adding novel \biokgs{} would require researchers and other users to contribute to \dataset{}'s codebase.

\section*{Ethical Considerations} 

\paragraph{Data Licensing and Access} Some datasets (like MIMIC-IV EL Challenge and ShARe/CLEF) have protective DUAs that do not allow redistribution. For these datasets, we provide annotation and mapping scripts, which can be run locally, under the assumption that the user has lawfully obtained the requisite raw data. \paragraph{Privacy and De-identification} Discharge summaries, clinical notes as well as social media posts were the only patient-facing corpora utilized in this study. They were released publicly in a de-identified format and cannot be re-identified through our processing methods.  
\paragraph{Ontology licensing and distribution} Controlled vocabularies subject to license restrictions are not redistributed, and thus scripts are provided which extract relevant paths and metadata, provided there exists a local install or relevant ontology sources.  
\paragraph{Potential Biases} The corpus inherits biases from constituent datasets, including geographic bias from US-centric hospital systems and linguistic bias from the predominantly English corpus. These factors should be taken under consideration when interpreting model performance and deploying systems built from derived models.

\section*{Acknowledgement}
The computational resources used were financed by the NWO research programme Computing Time on National Computer Facilities (grant 2024.015).

\bibliography{custom}
\appendix
\section{Source Corpora and Schema}
\label{sec:Dataset_appendix}

\subsection{Source Corpora Details}
Below, we provide details about the nine expert-annotated corpora that constitute \dataset{}.

\begin{figure*}[t]
\centering
\includegraphics[width=\textwidth]{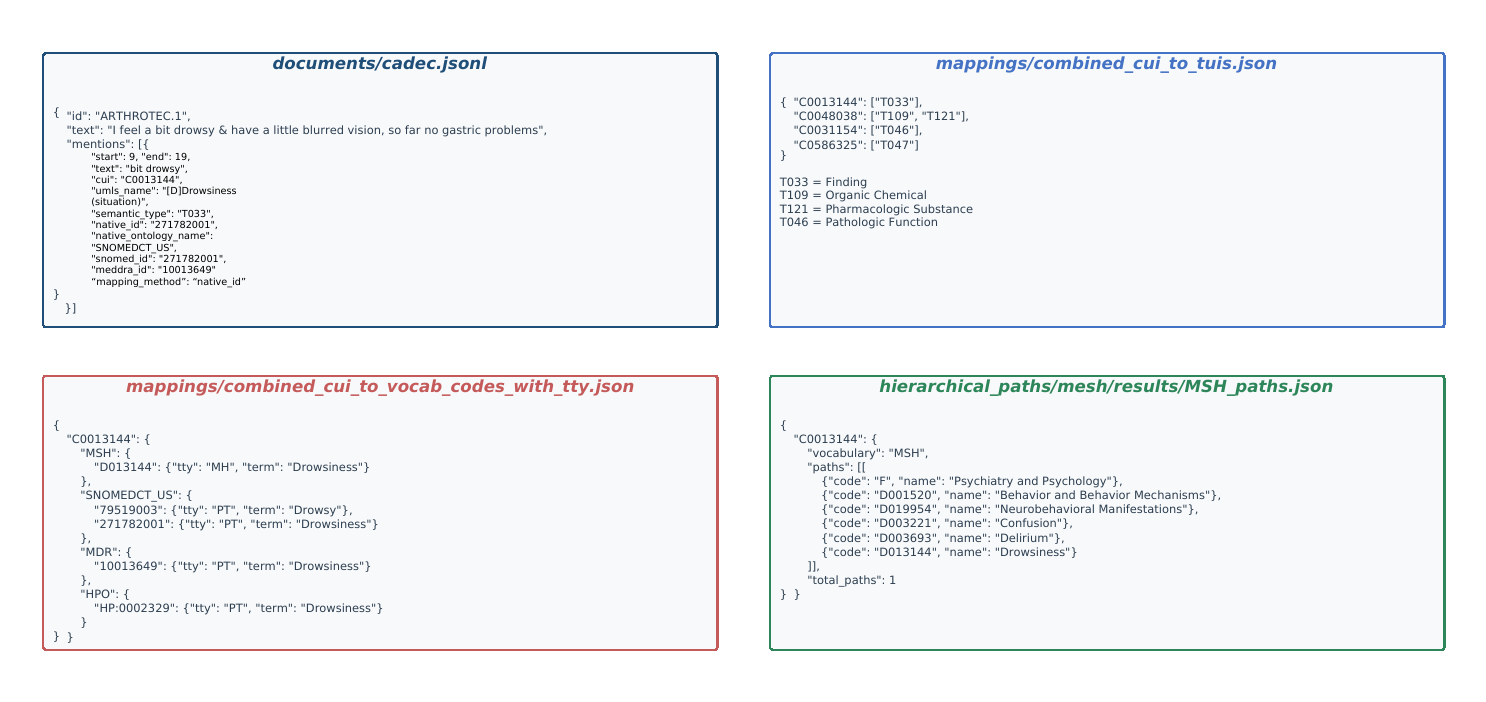}
\caption{An example showing the schema of the proposed dataset, which shows the four components (clockwise): (i) the preprocessed document data with annotations and CUI mappings, (ii) the semantic type mapping, (iii) the cross-vocabulary mappings, and (iv) the hierarchical ontological paths.}
\label{fig:json-schema}
\end{figure*}

\paragraph{MIMIC-IV SNOMED EL Challenge 2023 \cite{Davidson2025-kj}}
Includes 300 de-identified ICU discharge summaries richly annotated by two clinical experts with SNOMED CT disorder, procedure, drug, and device codes. Provides the largest publicly available gold-standard clinical EL dataset.
\paragraph{ShAReCLEF 2013 \cite{shareclef2013}} Part of an eHealth evaluation shared task, contains 199 hospital notes from Beth Israel hospital, double-annotated by clinical trainees and adjudicated by a senior MD for disorders, procedures, medications, and devices with UMLS CUIs.
\paragraph{Mantra GSC English \cite{10.1093/jamia/ocv037}} A multilingual dataset with 1,050 snippets drawn from patents, EU drug labels, and PubMed abstracts, covering 16 UMLS semantic groups annotated by three biomedical linguists.
\paragraph{BC5CDR \cite{b5cdr}} A popular BioNLP benchmark dataset, it contains 1,500 PubMed abstracts with exhaustive Chemical and Disease spans normalised to MeSH, manually annotated by a team of 3 biocurators. 
\paragraph{NCBI Disease \cite{Dogan2014-cu}} A relatively smaller dataset with 793 abstracts focusing exclusively on diseases, mapped to MeSH 2012 tree numbers. It was annotated by three biology graduate students.
\paragraph{MedMentions \cite{mohan2019medmentionslargebiomedicalcorpus}} The biggest dataset in \dataset{} in terms of scale, contains 4,392 PubMed abstracts with mentions linked to any of 3.2 million UMLS 2017AB concepts with no type restrictions. As the broadest coverage literature corpus, it provides scalability and long-tail concept retrieval. It was triple-annotated by seven life science graduate students. 
\paragraph{TAC 2017 ADR \cite{tac2017adr}} One of the most mention-dense and token-rich datasets in \dataset{}, it has 200 FDA Structured-Product Labels annotated for adverse-reaction spans (MedDRA linked) and drug names (RxNorm linked). It was also manually annotated by two pharma-safety scientists and reviewed by NIST.
\paragraph{CADEC \cite{Karimi2015-we}} One of the datasets from the social media/free text domain. It contains 1,250 Ask-a-Patient forum posts, labelled for patient-reported ADEs and drugs, and normalised to MedDRA, annotated in two steps by nurses and a biomedical ontologist.
\paragraph{COMETA \cite{basaldella-etal-2020-cometa}} Social media-based dataset with 20,000 Reddit/Twitter posts with symptom spans mapped to SNOMED CT 2019 version. It was annotated by five trained crowdsourced annotators and adjudicated by an MD.

\subsection{Dataset Schema}
\label{sec:dataset_schema}
To ensure both ease of use and computational efficiency, \dataset{} is distributed across several files, each with a distinct purpose. The primary data is provided in a standardized JSON format, containing the source documents and a list of all annotated mentions with their character offsets, original concept IDs, and canonical UMLS CUIs.
Supplementary annotations are provided in separate, optimized formats. Mappings from each unique CUI to its corresponding Semantic Type (TUI) and its parallel codes in other vocabularies are stored in simple key-value files. The core hierarchical annotations are structured as per-vocabulary lists of linear \textit{Root → Leaf} chains, each containing the codes and names for each concept along the path. 
Figure \ref{fig:json-schema} demonstrates the final schema of the dataset.

\section{Additional Data Analysis}

In this section, we present further analyses. 
\subsection{Document length}
\begin{figure}[t]
  \centering
  \includegraphics[width=\linewidth]{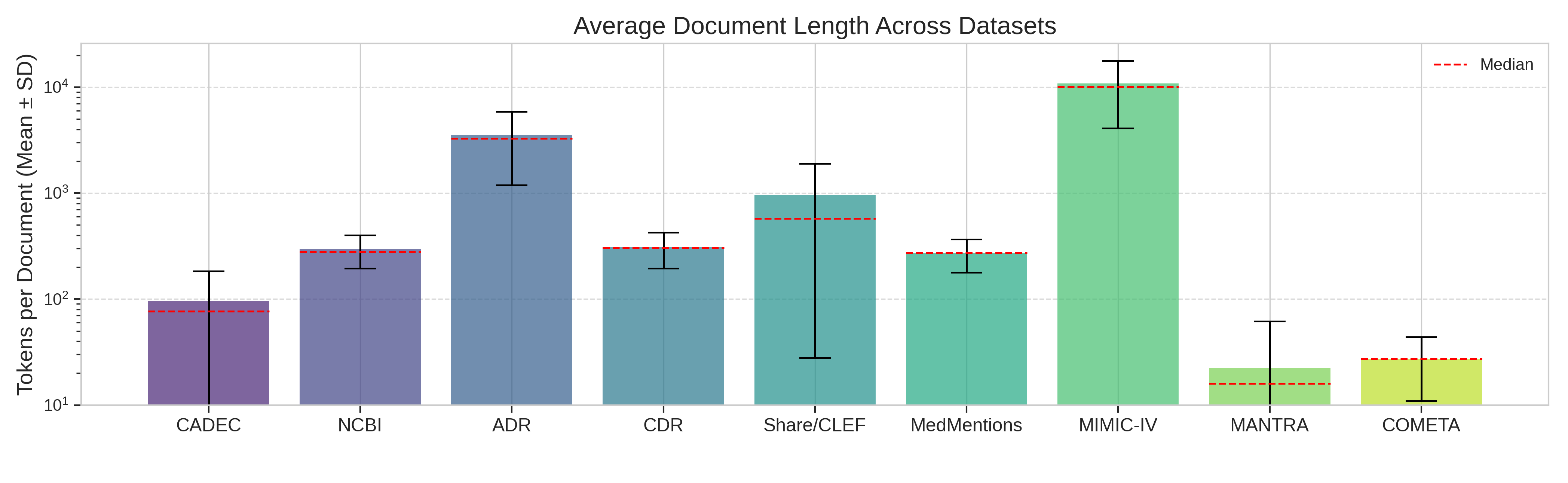}
  \caption{The mean and median document length for each dataset, shown in terms of BERT tokens.}
  \label{fig:doc_length}
\end{figure}

As illustrated in Figure \ref{fig:doc_length}, the document lengths vary considerably across the source corpora. Clinical notes (MIMIC-IV) and drug labels (ADR) feature the longest documents, whereas snippets from patents and abstracts (Mantra-GSC) and social media posts (COMETA) are significantly shorter.

\subsection{Semantic type distribution}
\begin{figure*}[t]
    \centering
    \includegraphics[width=\textwidth]{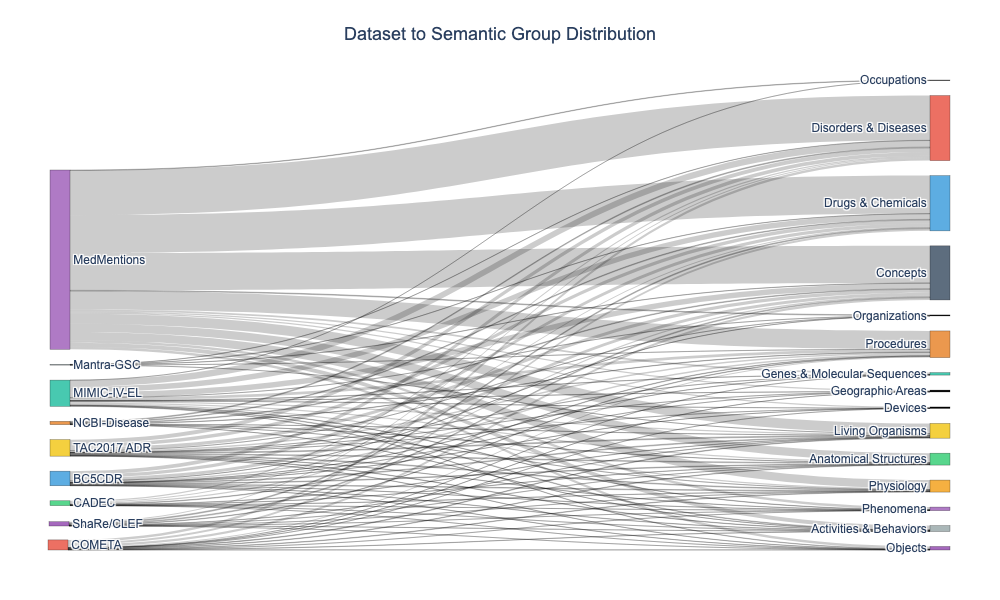}
    \caption{Contribution of each source dataset to concepts belonging to 15 major semantic type categories.}
    \label{fig:sankey}
\end{figure*}

To visualize the contribution of each source dataset to the overall semantic diversity, we present a Sankey diagram in Figure \ref{fig:sankey}. This plot depicts the flow of mentions from each source dataset to the 15 most frequent semantic type categories, confirming that MedMentions provides the broadest coverage across all categories.
\subsection{Ontology Path Statistics}
\begin{figure}[t]
    \centering
    \includegraphics[width=\linewidth]{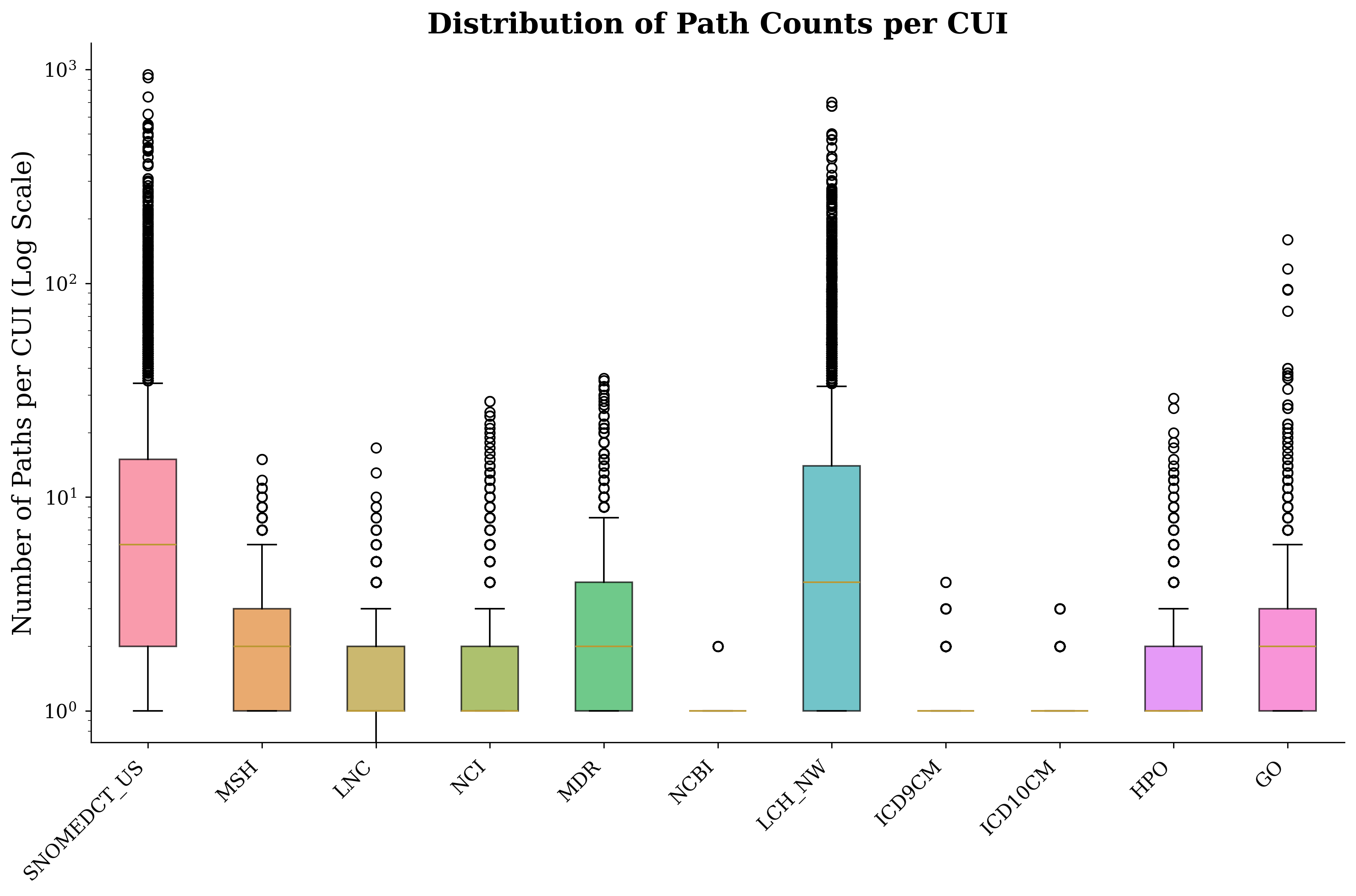}
    \caption{Number of paths per CUI in each vocabulary.}
    \label{fig:num_paths}
\end{figure}

\paragraph{Number of Paths} The number of hierarchical paths per concept differs significantly across vocabularies, as shown in the boxplots in Figure \ref{fig:num_paths}. SNOMED CT and LCH\_NW exhibit the largest variance, whereas vocabularies such as NCBI and ICD are largely monohierarchical. 

\begin{figure}[t]
    \centering
    \includegraphics[width=\linewidth]{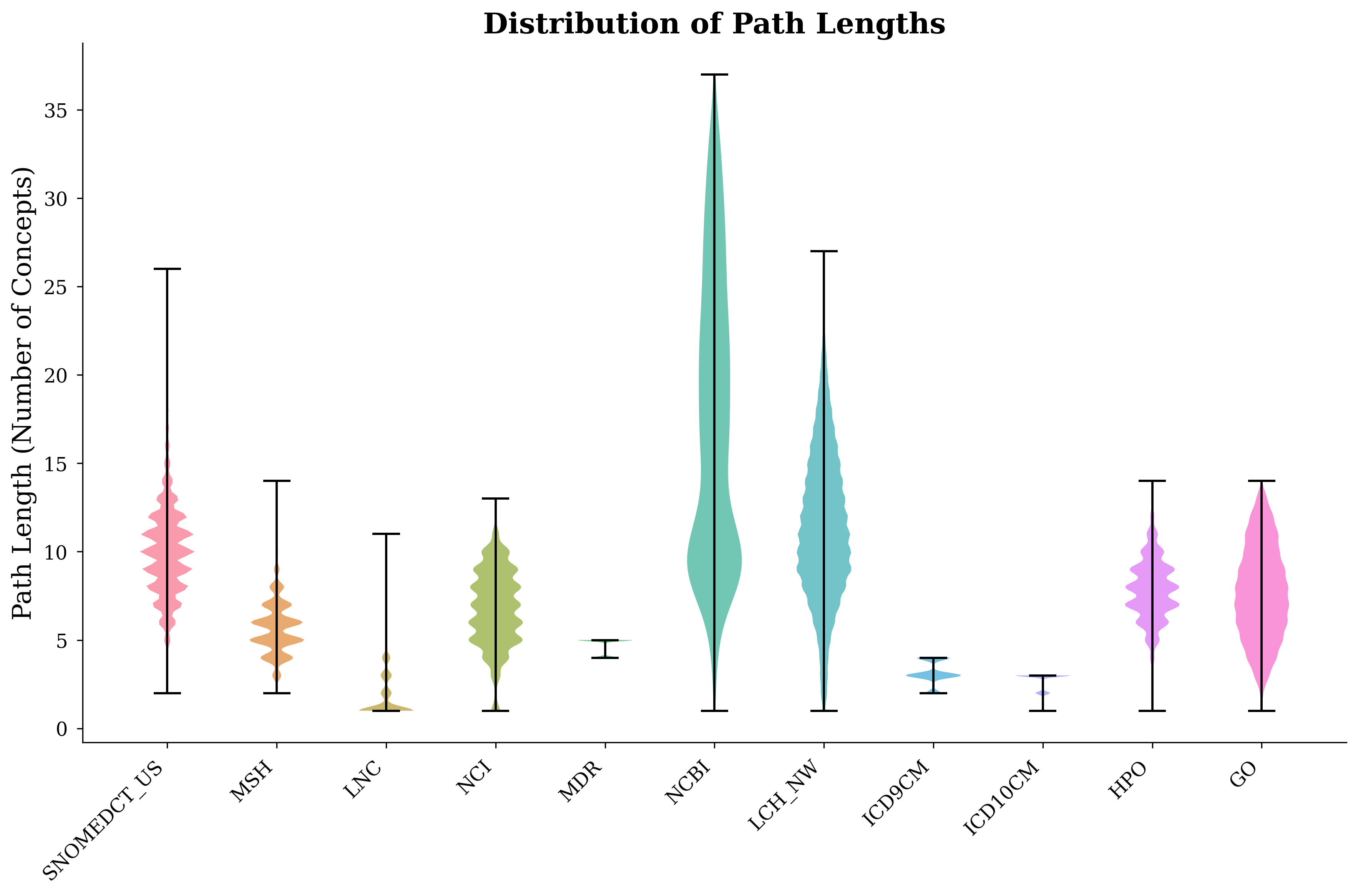}
    \caption{Path length distribution per vocabulary.}
    \label{fig:paths_violin}
\end{figure}

\paragraph{Path Length Distribution}
The violin plot in Figure \ref{fig:paths_violin} illustrates the distribution of path lengths. On average, NCBI has the deepest paths, while MedDRA shows the least variance, consistent with its well-defined five-level hierarchy.

\section{Benchmark and Baseline Experiments}
\label{sec:baseline_appendix}
This section provides additional details on our experimental setup and preliminary results.

\subsection{Named Entity Recognition (NER) Setup}
\label{sec:ner_setup}
 We cast NER as token-level sequence labeling with a BIO scheme over chunks of our documents. Mentions are judged correct only if span boundaries and types match exactly.

\paragraph{Metrics} We calculate strict and lenient micro-F1 scores per class and overall.
We also calculate span-detection performance regardless of the predicted entity type. 

\paragraph{Data Preprocessing} First, a document was segmented into chunks of 512 characters with a 128-character sliding window. Mention offsets were recalculated relative to each chunk. Source domain and dataset information were preserved to facilitate ablation studies. An example of the final JSON format is shown below:

{\small
\begin{verbatim}
  {
    "chunk_id": "227508_0",
    "source_dataset": "cdr",
    "source_domain": "abstracts",
    "text": "Naloxone reverses the 
    antihypertensive effect of clonidine...",
    "entities": [
      {
        "start": 0,
        "end": 8,
        "label": "CHEM",
        "text": "Naloxone",
        "cui": "C0027358",
        "original_start": 0,
        "original_end": 8
      },
    ],
    "entity_types": ["CHEM", "DISO", "MISC"],
    "chunk_start": 0,
    "chunk_end": 512,
    "doc_length": 1135
  }
\end{verbatim}
}

\paragraph{Data Split and Labels} For datasets with pre-defined splits, we retained them. For those without, we created a 50/10/40 train/dev/test split. If only a train/test split existed, a 10\% dev set was carved out from the training data. Models were trained on 11 high-level semantic type classes derived from UMLS Semantic Groups (see Figure \ref{fig:sem_type}). 
\paragraph{Experiment Paradigms} To demonstrate robustness, get insights into the dataset composition, and highlight the value of cross-domain unification, we run four types of experiments.
\begin{itemize}
    \item \textbf{Full-Mix}: Train on the union of all training splits; evaluate on the union of all test splits.
    \item \textbf{In-Domain}: Build train/dev/test splits per single dataset.
    \item \textbf{Leave-One-Dataset-Out (LODatO)}: Hold out one dataset for testing, and train on all other datasets. 
    \item \textbf{Leave-One-Domain-Out (LODomO)}: Hold out all datasets from one domain for testing (clinical / literature / social / label), training on all datasets from the remaining domains. 
\end{itemize}

\paragraph{Models} For the \textbf{Full-Mix} setting, we fine-tune and evaluate five biomedical PLMs pretrained on different domains. These models are listed below:
\begin{itemize}[nosep]
  \item \textbf{GatorTron-base}: An encoder-only transformer model pre-trained on a large corpus of over 82 billion words from de-identified clinical notes and clinical trial publications, developed by the University of Florida~\cite{yang2022gatortron}.
  \item \textbf{ClinicalBERT}: A BERT model pre-trained on the MIMIC-III dataset, which contains de-identified health records, making it highly specialized for tasks on clinical notes~\cite{huang2019clinicalbert}.
  \item \textbf{PubMedBERT}: A BERT model pre-trained from scratch exclusively on biomedical literature, specifically 21GB of text from PubMed abstracts and full-text articles~\cite{pubmedbert}.
  \item \textbf{BioBERT}: One of the first domain-specific BERT models, initialized from Google's BERT and continually pre-trained on a large-scale biomedical corpus including PubMed abstracts and PMC full-text articles~\cite{lee_biobert_2019}.
  \item \textbf{BlueBERT}:  BERT model pre-trained on a combination of biomedical (PubMed abstracts) and clinical data (MIMIC-III notes), designed to perform well on a diverse range of biomedical and clinical NLP tasks~\cite{peng2019bluebert}.
  \item \textbf{GliNER-BioMed} We also evaluated GliNER-BioMed \cite{yazdani2025gliner}, a task-specific NER model in a zero-shot setting. GliNER is a generative encoder-decoder model that takes natural-language class labels, along with the input sentence, and outputs spans of mentions belonging to those classes. For our evaluation, we used all variations of our semantic classes in natural language as potential class names to pass to GliNER. E.g., for Disorder (DISO), we passed \{`disease', `disorder', `condition', `syndrome', `pathology', `findings'\} along with sentences, and then mapped the extracted entities to our class names for consistent comparison.

\end{itemize}

For all \textbf{ablation studies} (LODatO, LODomO), we use \textbf{PubMedBERT} because it is consistently strong across domains and classes, yet faster and lighter than GatorTron. This keeps computation manageable and isolates the effect of the data splits from that of the model size. 
\paragraph{Hyperparameter Tuning} For all the experiments that involved fine-tuning models: full mix, and the various ablations as described in \ref{sec:ner_setup}, we implemented a thorough hyperparameter tuning across a range of hyperparams (focal loss, crf layer, batch size, learning rate, weight decay, warmup ratio, early stopping) through randomized trials. The best model in a full-mix setting, i.e., GatorTron-base, achieved the best performance with a linear-CRF layer, a base learning rate of 5e-5 with a CRF layer learning rate of 1e-5, a batch size of 32, 0.01 weight decay, and 0.1 warmup. Additionally, we used a 3x random oversampling to balance underrepresented classes, along with a class-weighted loss function.

\subsubsection{Preliminary NER Experiment Results and Discussion}

Table \ref{tab:ner_main} shows the overall performance of the NER models on our unified dataset. GatorTron performed best across all metrics, with PubmedBERT a close second. GliNER zero-shot performed poorly across most categories, except Disorders and Chemicals/Drugs. Comprehensive results across classes, domains, and datasets, along with observations from the ablation studies, are presented below. 

\begin{table}[t!]
  \centering
  \small
  \setlength{\tabcolsep}{3pt}
  \resizebox{\columnwidth}{!}{%
    \begin{tabular}{lccc}
      \toprule
      \textbf{Model} & \textbf{Strict $F_{1}$} &
      \textbf{Lenient $F_{1}$} & \textbf{EA $F_{1}$}\\
      \midrule
      \multicolumn{4}{l}{\textbf{Finetuned Models}} \\
      GatorTron-base & \textbf{0.663} & \textbf{0.746} & \textbf{0.760}\\
      PubMedBERT     & 0.642 & 0.728 & 0.743\\
      ClinicalBERT   & 0.615 & 0.713 & 0.726\\
      BioBERT        & 0.623 & 0.717 & 0.730\\
      BlueBERT       & 0.606 & 0.705 & 0.720\\
      \midrule
      \multicolumn{4}{l}{\textbf{Zero-shot}} \\
      GLiNER-BioMed & 0.365 & 0.482 & 0.524\\
      \bottomrule
    \end{tabular}%
  }
  \caption{Micro-averaged NER test performance across all 11 semantic groups. \textbf{EA $F_1$}: Entity-agnostic $F_{1}$.}
  \label{tab:ner_main}
\end{table}

\begin{figure}[t]
    \centering
    \includegraphics[width=\linewidth, height=0.2\textheight]{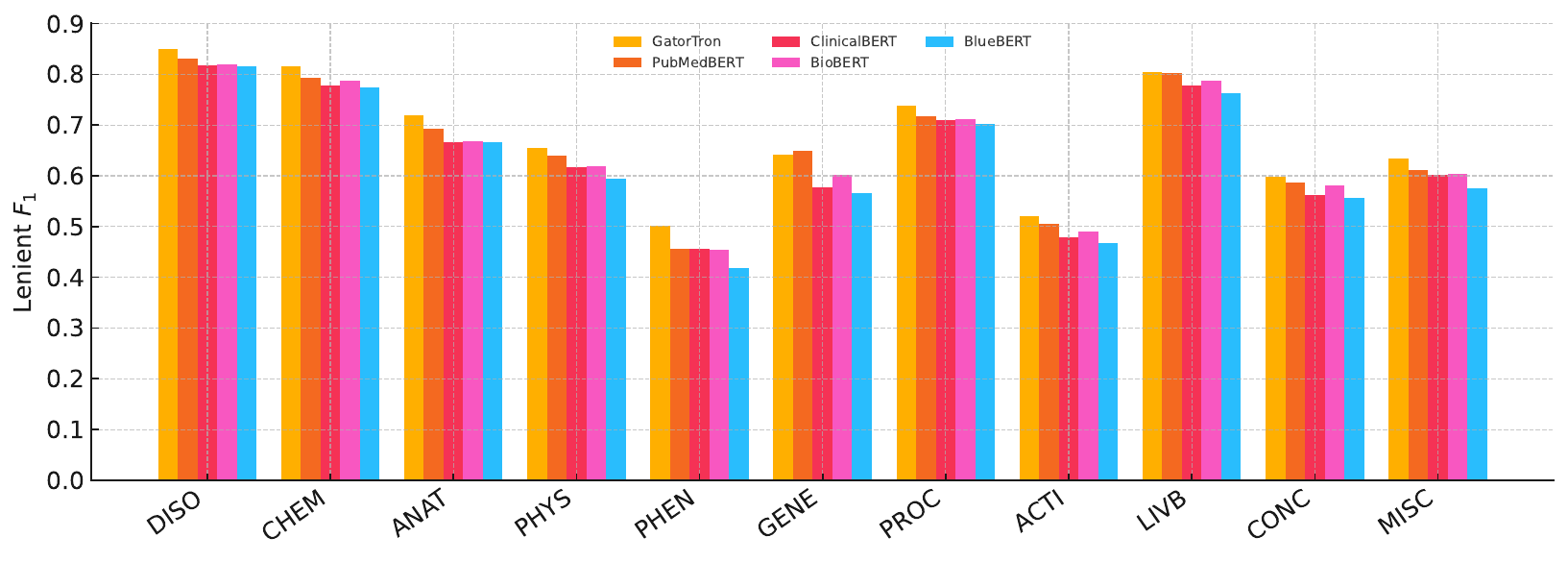}
    \caption{Performance of fine-tuned models in full mix setting across the 11 classes, shown using lenient F1.}
    \label{fig:lenient_f1_ner}
\end{figure}

\paragraph{Full Mix} Figure \ref{fig:lenient_f1_ner} shows the per-class F1 performance for the five fine-tuned models in the Full-Mix setting. 

\begin{figure}[t]
    \centering
    \includegraphics[width=\linewidth, height=0.2\textheight]{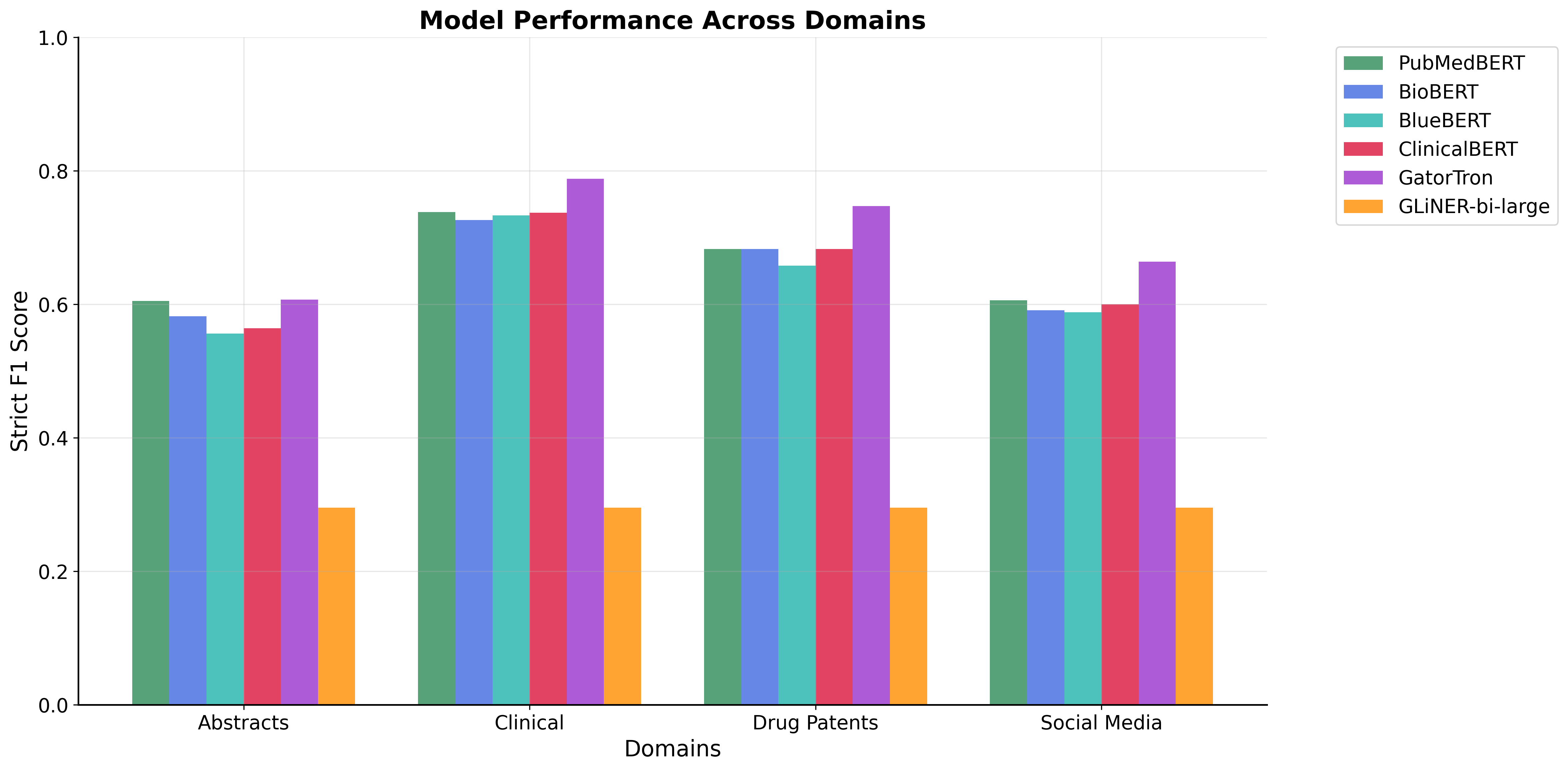}
    \caption{Performance of models across the four main domains, shown using strict F1.}
    \label{fig:strict_domain_ner}
\end{figure}

Figure \ref{fig:strict_domain_ner} breaks down the strict F1 performance by domain for all models, including the zero-shot GliNER-BioMed.

\begin{figure}[!t]
    \centering
    \includegraphics[width=\linewidth]{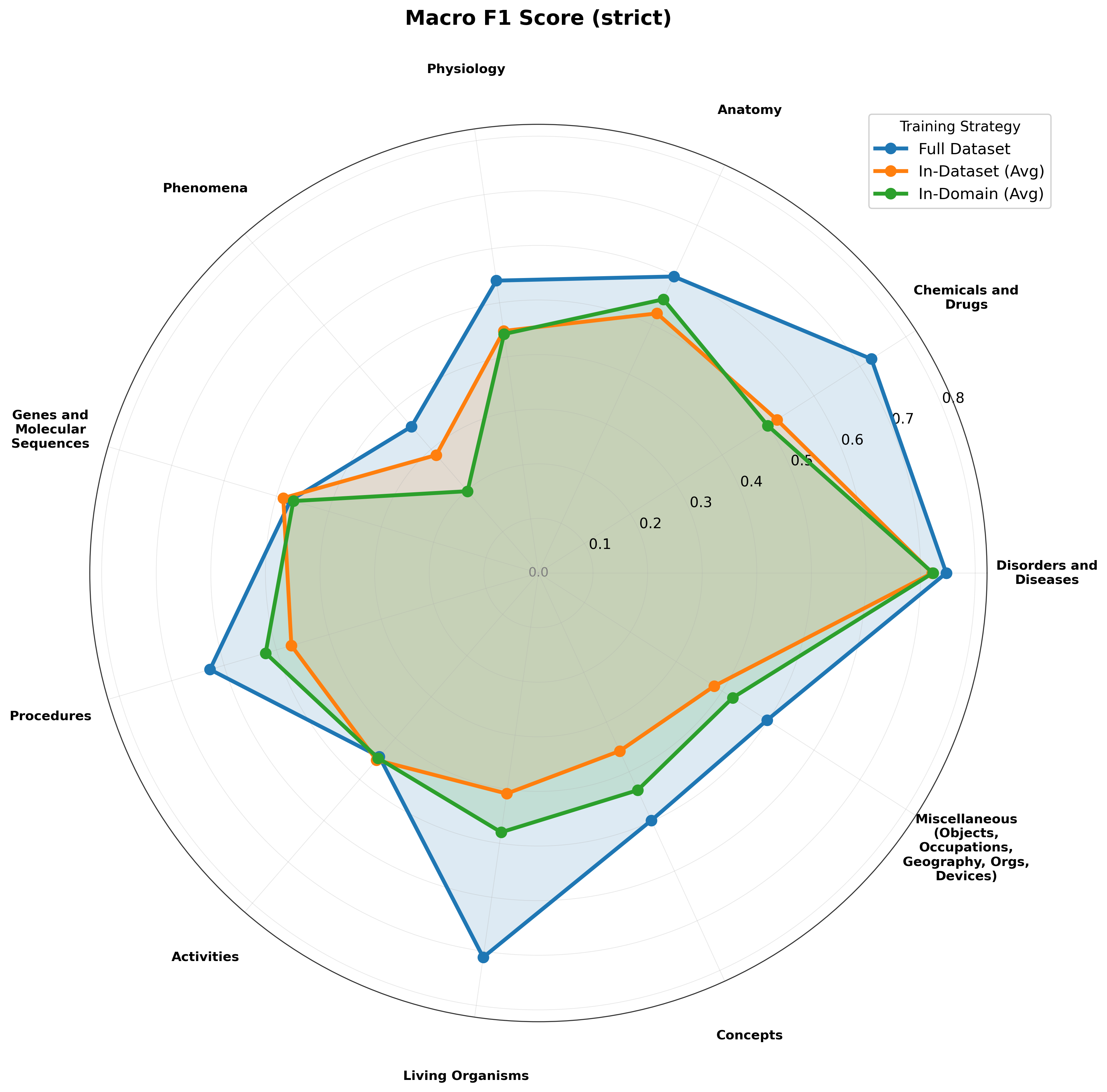}
    \caption{Figure showing the macro average performance over semantic types of the NER model in-domain, in-dataset, and a full mix setting}
    \label{fig:abation_in_domain}
\end{figure}

\begin{figure}[!t]
    \centering
    \includegraphics[width=\linewidth, height=0.25\textheight]{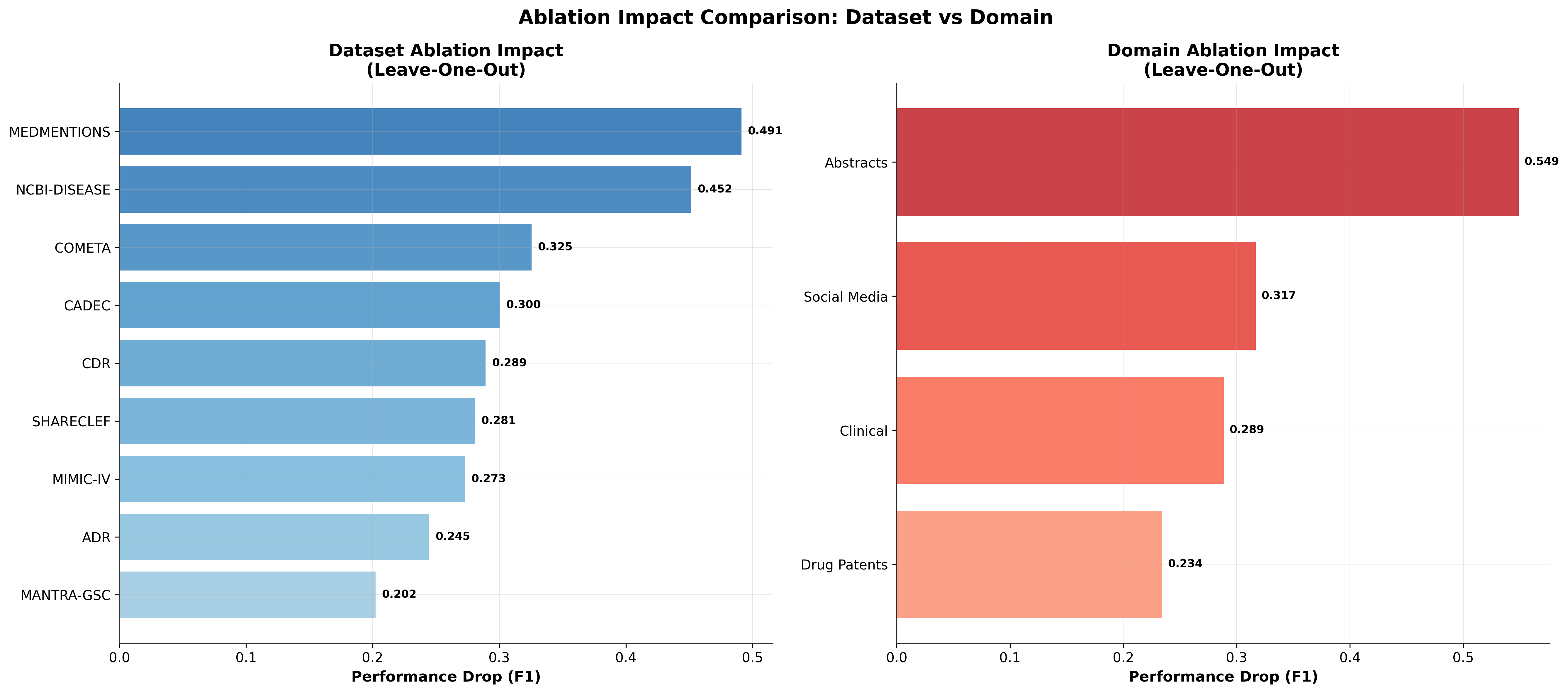}
    \caption{Figure showing the performance $\Delta$ in the LOO and LOODom ablation experiments}
    \label{fig:abation_loo}
\end{figure}

\paragraph{Ablations}
The radar chart in Figure \ref{fig:abation_in_domain} compares the average performance of models trained on the full mix versus those trained in-domain or in-dataset, demonstrating the clear benefit of \dataset{}. Figure \ref{fig:abation_loo} quantifies the performance impact of holding out each dataset and domain, with MedMentions and the abstracts domain showing the most significant impact due to their scale.



\subsection{Additional Entity Linking Results}

\begin{figure}[!t]
    \centering
    \includegraphics[width=\linewidth]{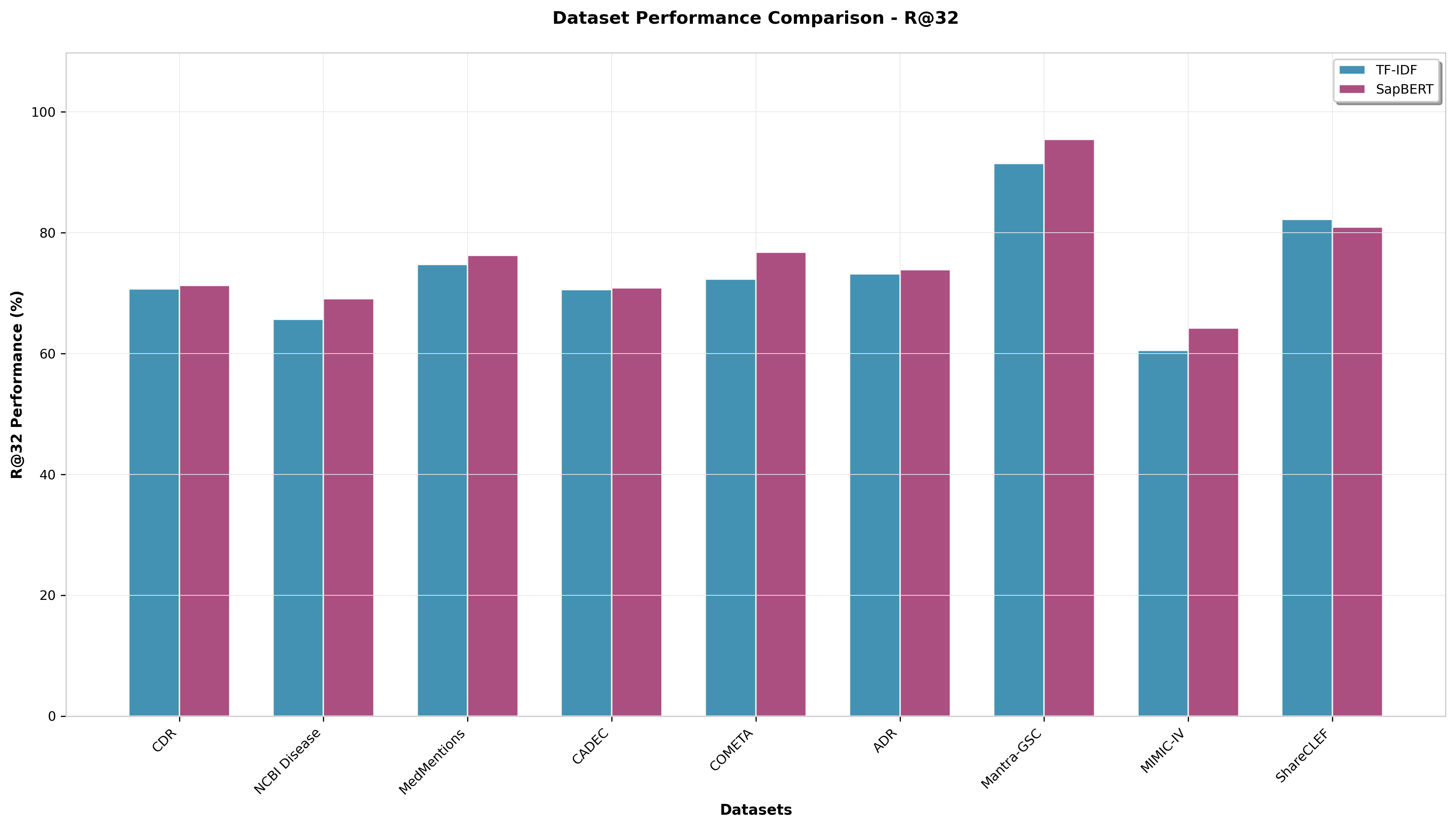}
    \caption{Performance of TF-IDF and SapBERT Candidate Generation across datasets}
    \label{fig:EL_dataset}
\end{figure}

Here, we present the performance of the TF-IDF and SapBERT-based entity linkers across various verticals. Figure \ref{fig:EL_dataset} shows the candidate-generation-based EL metrics for mentions across the nine datasets using the Acc@32 metric. We see that SapBERT consistently outperforms TF-IDF by small margins. Mantra-GSC, ShaRE/CLEF, and MedMentions show the best performance, which follows logically from the fact that their original ground truth annotations were in UMLS (Table~\ref{tab:harmony_core}), and the dictionary we created utilized UMLS concepts.

\begin{figure}[!t]
    \centering
    \includegraphics[width=\linewidth]{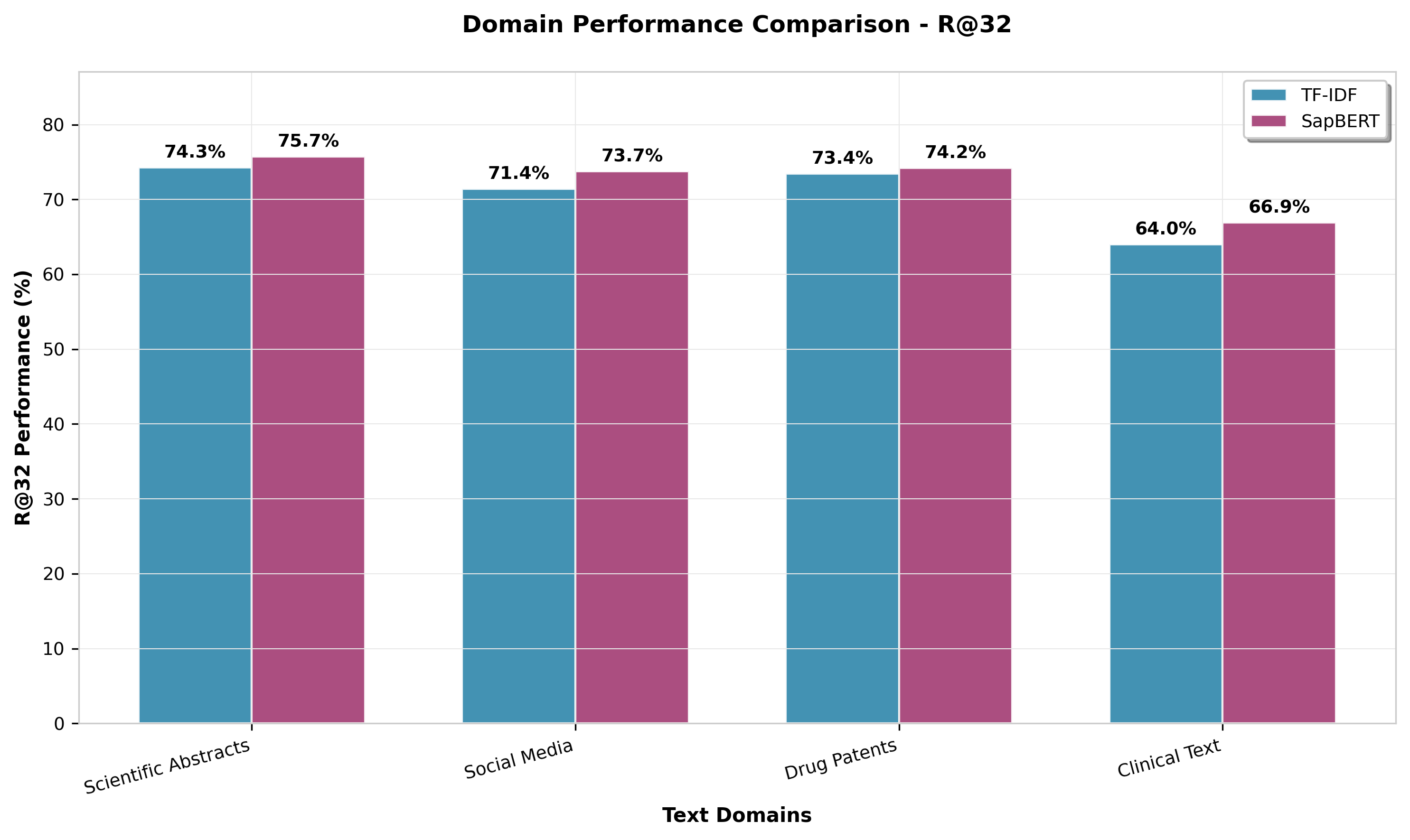}
    \caption{Performance of TF-IDF and SapBERT Candidate Generation across domains}
    \label{fig:EL_domain}
\end{figure}

Similarly, in Figure \ref{fig:EL_domain}, we visualize the EL performance across the four domains in our dataset. The major takeaway is the underwhelming performance on clinical notes, suggesting these terms have multiple surface forms or lack vocabulary integration.

Table \ref{tab:el_entity_type} details the performance of these two linkers over the various semantic groups. Living Beings and Activity mentions show the highest performance, while Procedure and MISC classes, including devices, occupation, and organization, show weak performance. 

\begin{table}[t]
\centering
\small
\setlength{\tabcolsep}{4pt}
\renewcommand{\arraystretch}{1.05}
\caption{Candidate Generation performance per Entity Type by SapBERT and TF-IDF, \textbf{boldface} denotes highest R@32 and \underline{underline} denotes highest R@1 per type}
\label{tab:el_entity_type}
\begin{tabularx}{\columnwidth}{l r *{4}{>{\centering\arraybackslash}X}}
\toprule
\multirow{2}{*}{Entity Type} & \multirow{2}{*}{Count} & \multicolumn{2}{c}{TF-IDF} & \multicolumn{2}{c}{SapBERT} \\
\cmidrule(lr){3-4} \cmidrule(lr){5-6}
& & R@1 & R@32 & R@1 & R@32 \\
\midrule
Disorder      & 61{,}054 & \underline{55.9\%} & 74.3\% & 52.9\% & \textbf{76.5\%} \\
Concept       & 32{,}094 & \underline{54.9\%} & 76.6\% & 51.9\% & \textbf{77.7\%} \\
Procedure     & 20{,}528 & 27.0\% & 58.2\% & \underline{28.9\%} & \textbf{60.3\%} \\
Chemical      & 19{,}216 & \underline{55.4\%} & 69.5\% & 49.2\% & \textbf{70.2\%} \\
Living Being  & 8{,}653  & \underline{62.6\%} & 81.0\% & 56.6\% & \textbf{82.5\%} \\
Anatomy       & 7{,}502  & \underline{53.6\%} & 73.8\% & 42.7\% & \textbf{75.6\%} \\
Physiology    & 7{,}175  & \underline{51.9\%} & 75.2\% & 44.6\% & \textbf{75.7\%} \\
Miscellaneous & 5{,}045  & \underline{41.3\%} & 59.7\% & 38.3\% & \textbf{62.3\%} \\
Activity      & 3{,}751  & 56.4\% & 78.9\% & \underline{59.1\%} & \textbf{79.4\%} \\
Phenomenon    & 2{,}009  & \underline{40.6\%} & 63.3\% & 38.9\% & \textbf{64.9\%} \\
Gene          & 1{,}523  & \underline{42.5\%} & 64.5\% & 34.9\% & \textbf{66.6\%} \\
\bottomrule
\end{tabularx}
\end{table}

\begin{figure*}[t!]
  \centering
  \setlength{\abovecaptionskip}{4pt}
  \setlength{\belowcaptionskip}{0pt}
  \begin{adjustbox}{max totalsize={\textwidth}{0.95\textheight},center}
    \begin{tabular}{@{}cc@{}}
      \includegraphics[width=0.48\textwidth]{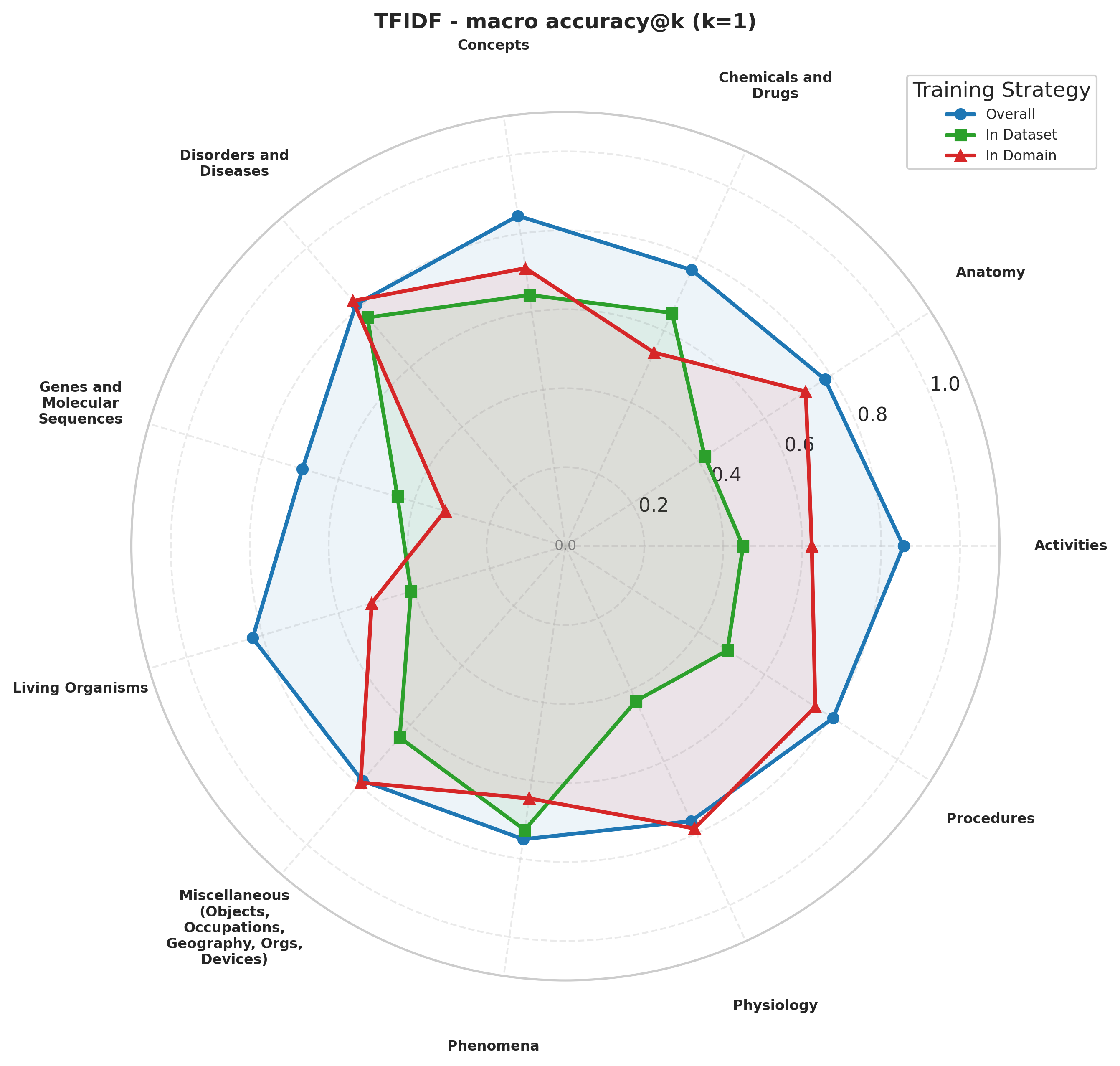} &
      \includegraphics[width=0.48\textwidth]{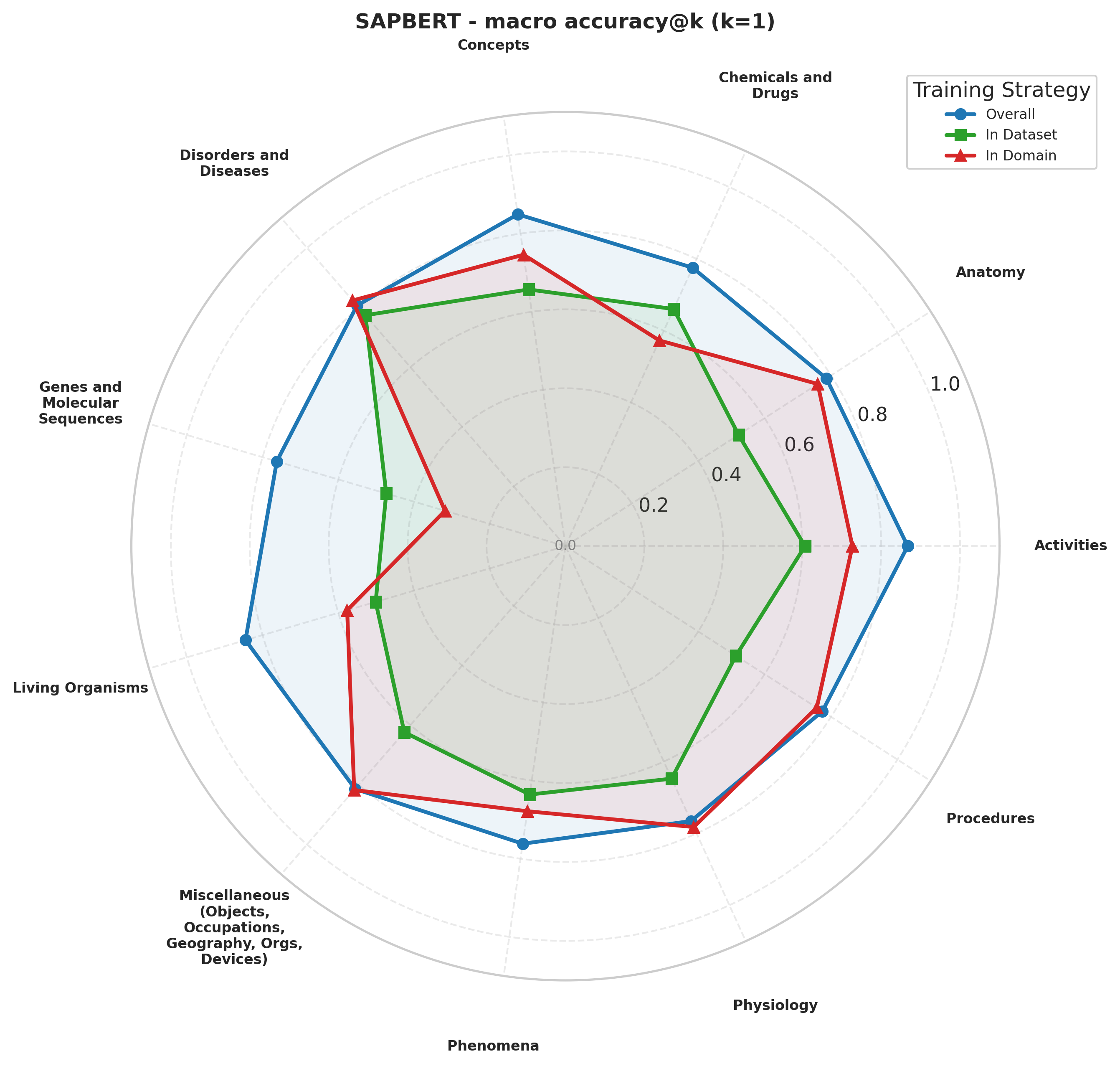} \\
      \includegraphics[width=0.48\textwidth]{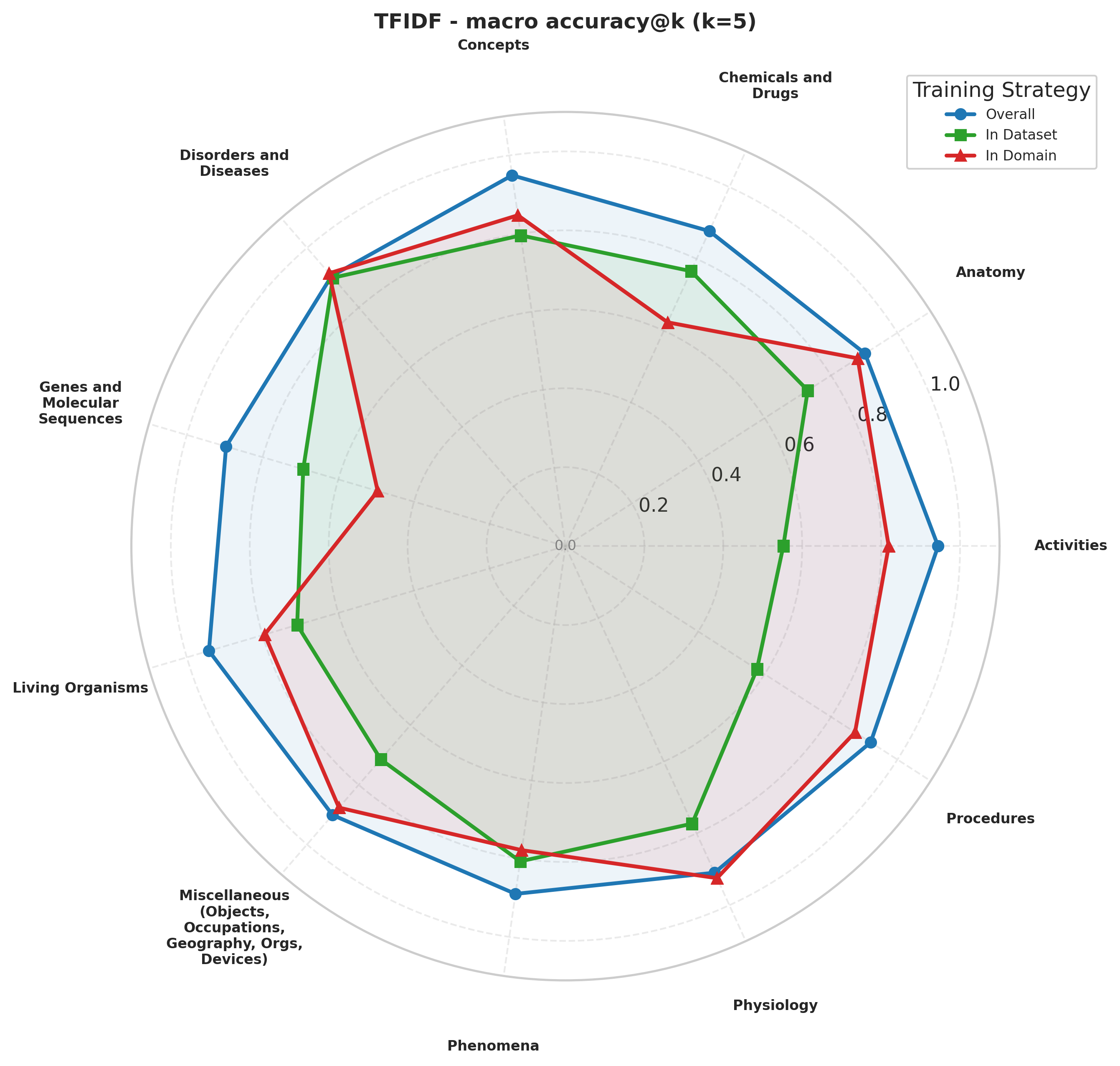} &
      \includegraphics[width=0.48\textwidth]{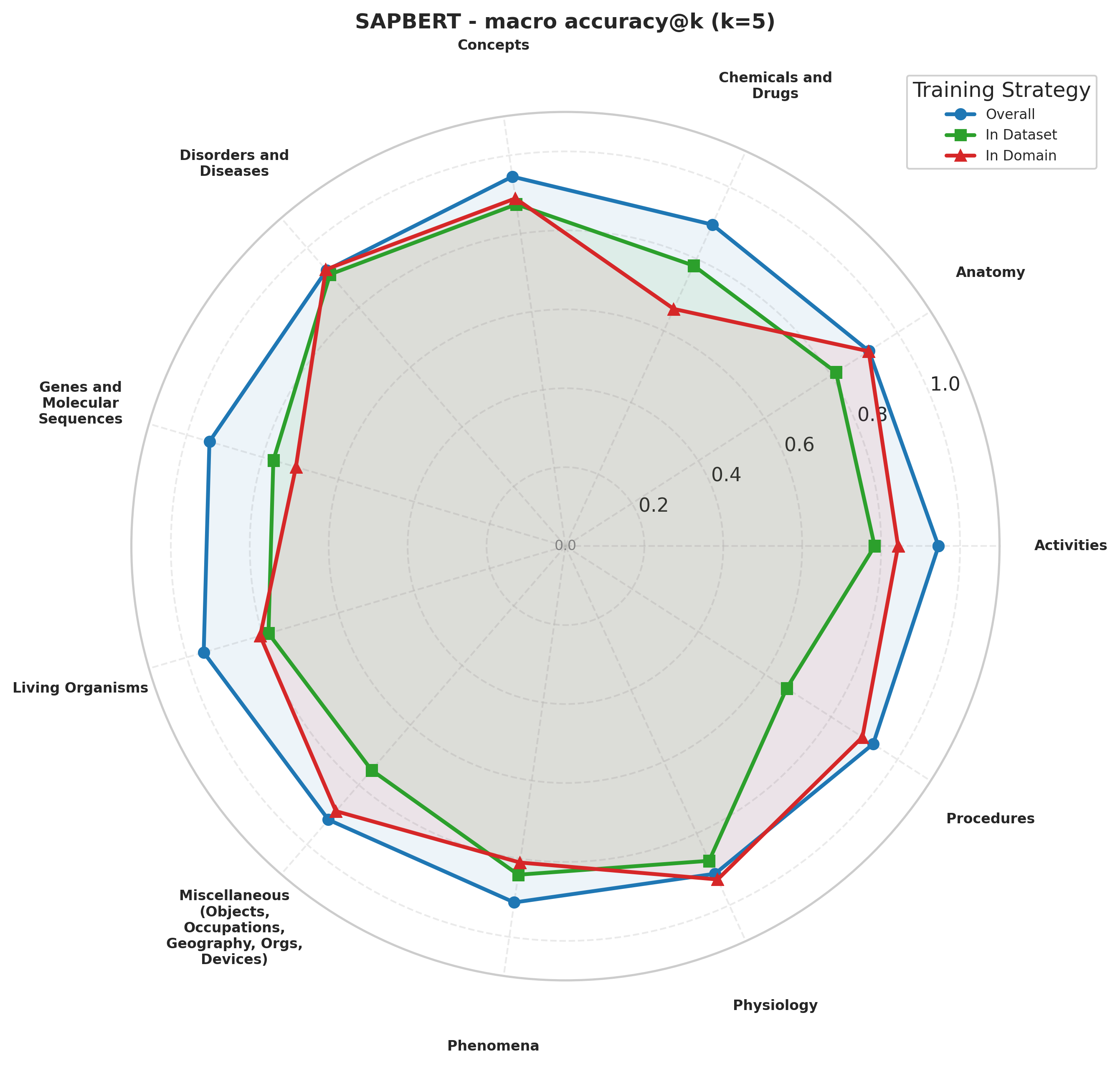} \\
      \includegraphics[width=0.48\textwidth]{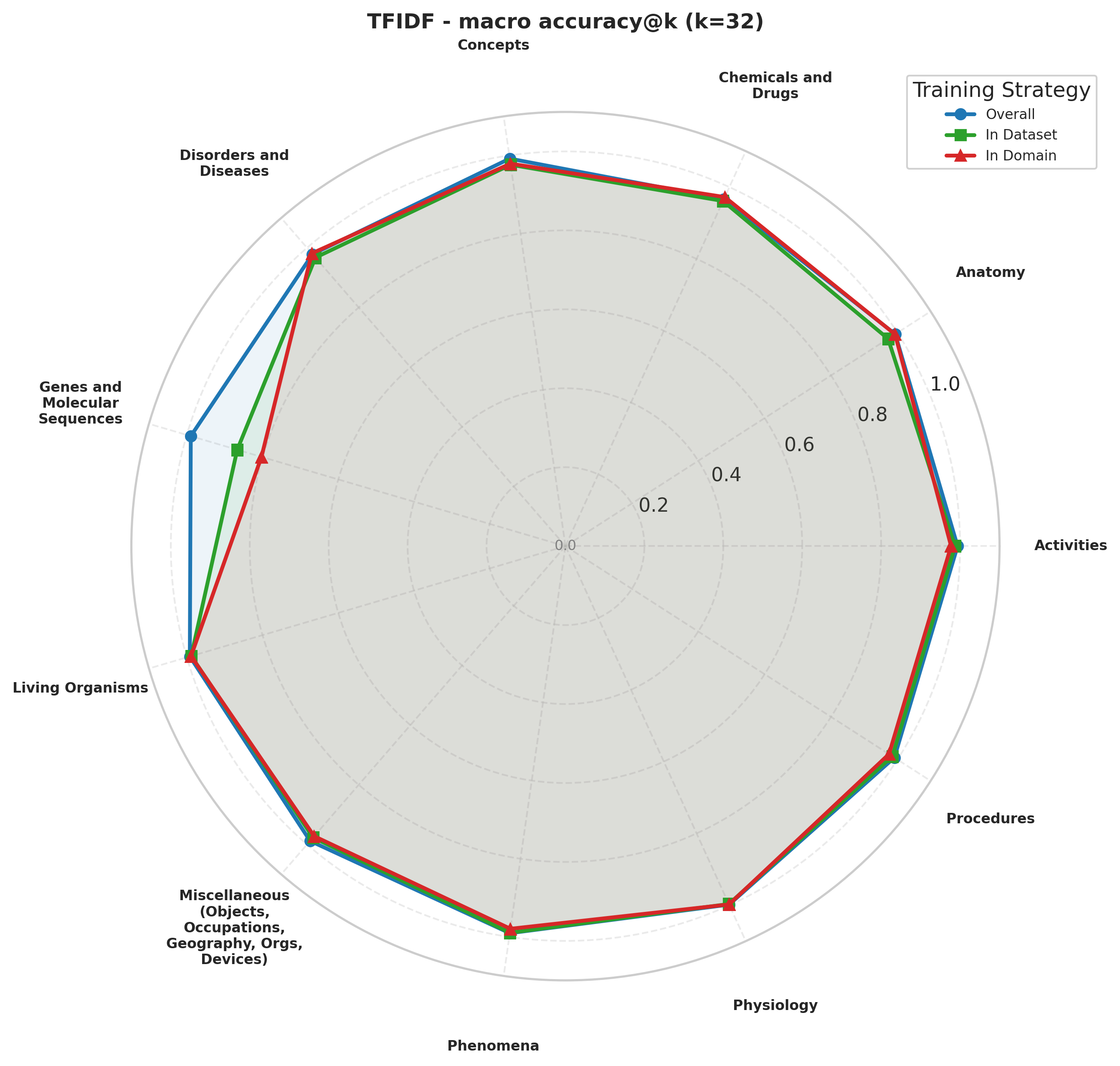} &
      \includegraphics[width=0.48\textwidth]{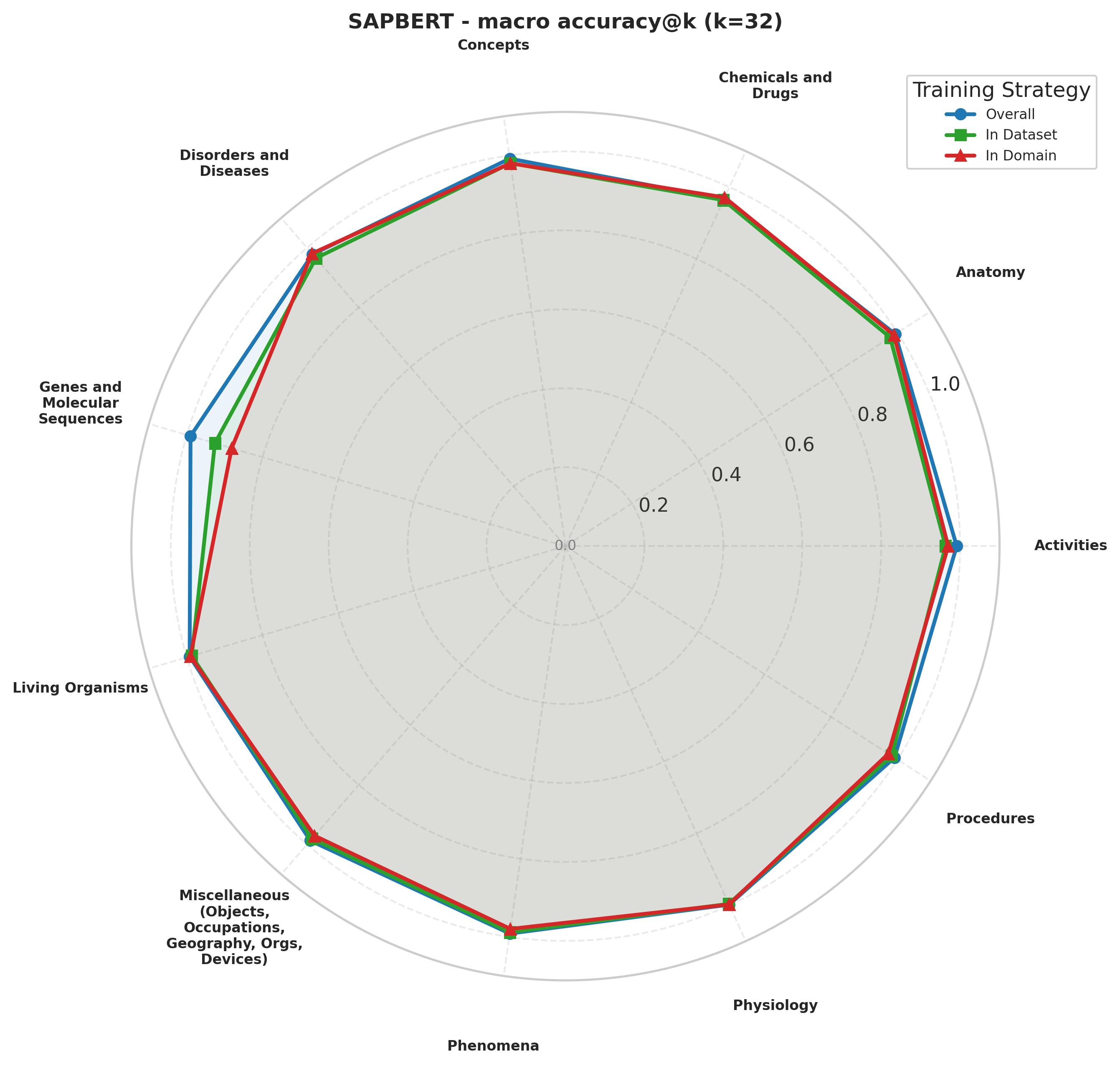} \\
      \includegraphics[width=0.48\textwidth]{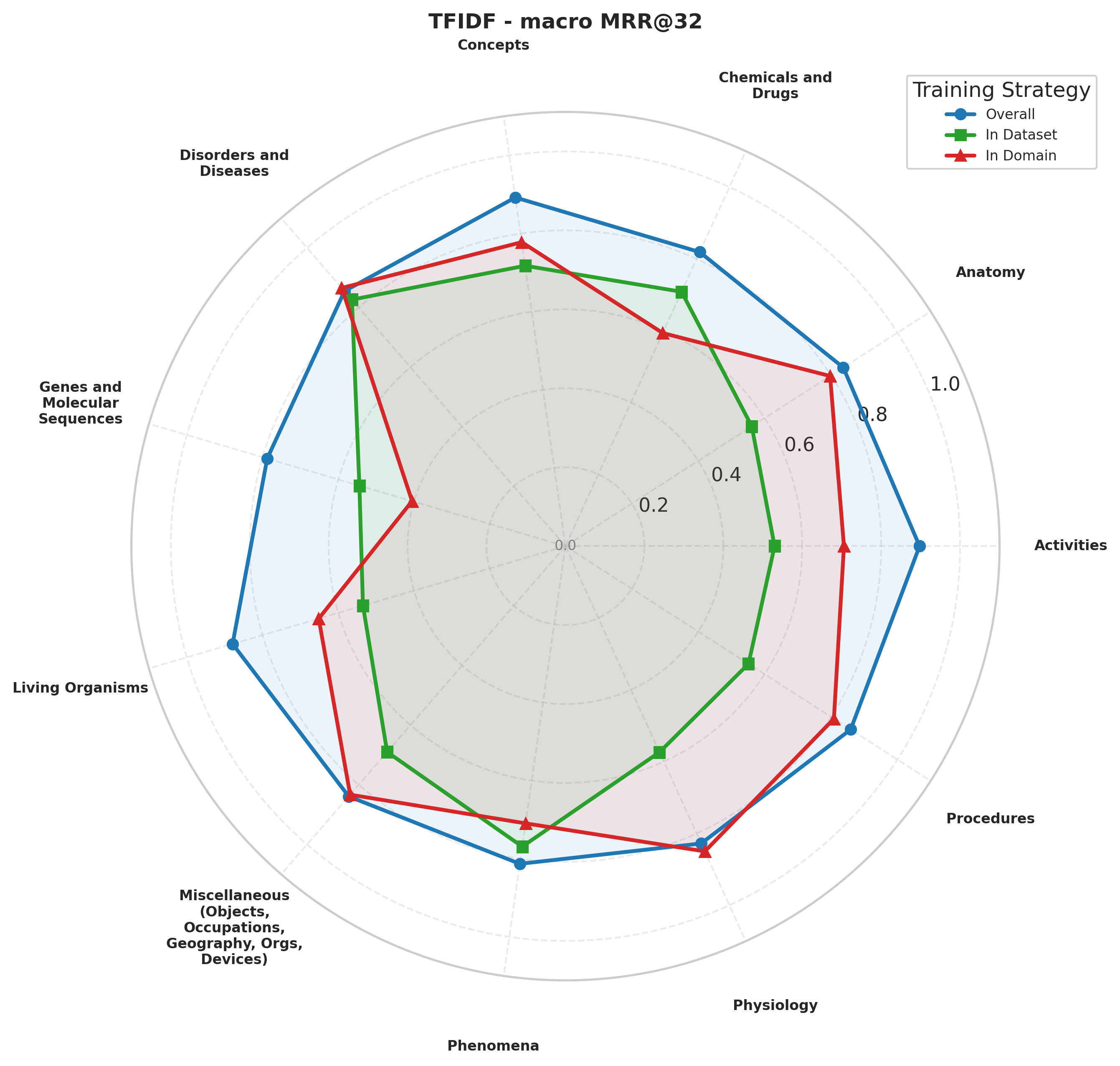} &
      \includegraphics[width=0.48\textwidth]{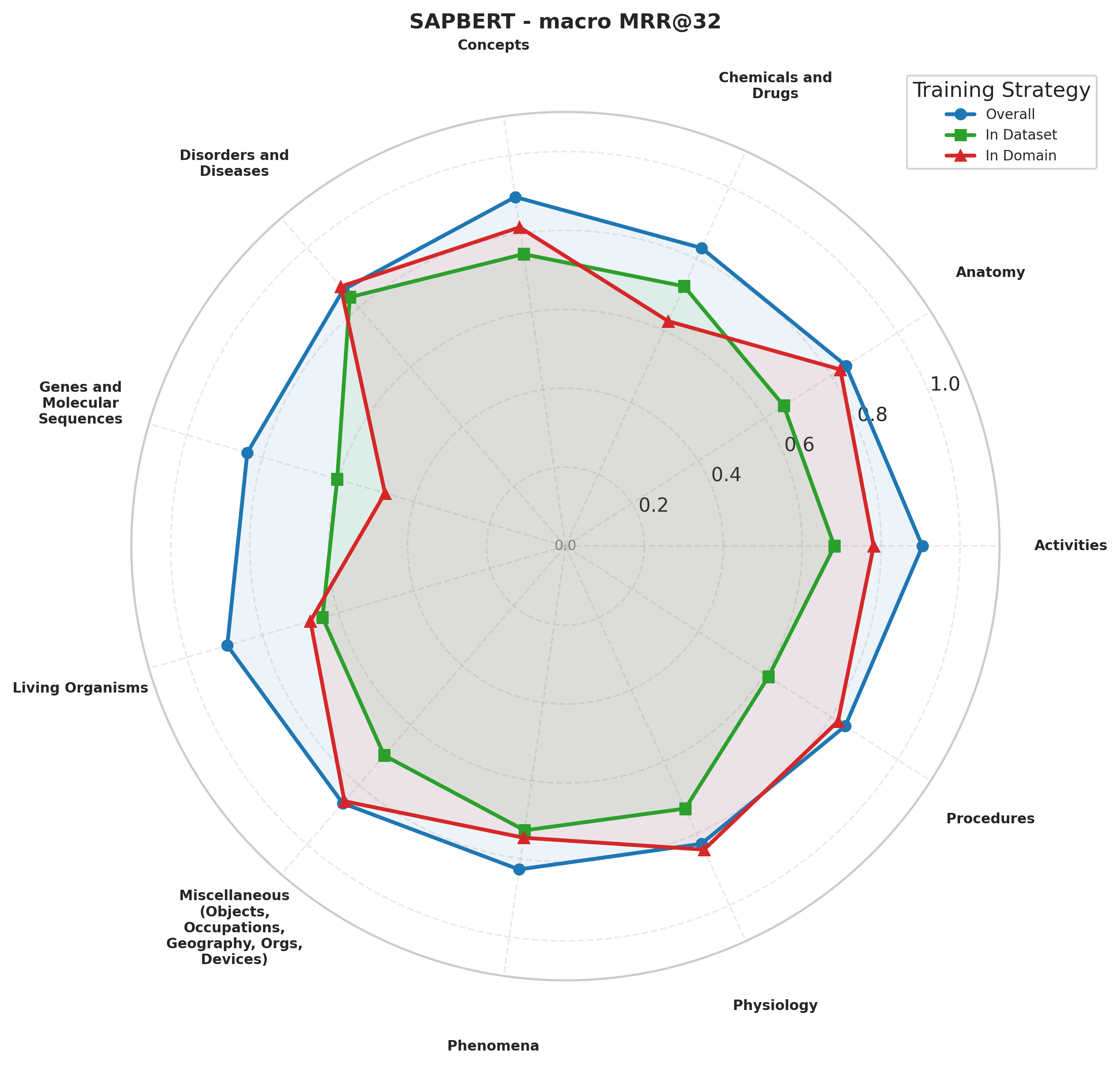} \\
    \end{tabular}
  \end{adjustbox}
  \caption{Entity linking performance across metrics (rows: Acc@1, Acc@5, Acc@32, MRR@32), averaged over all semantic-type , for all three training strategies. Left: TF-IDF+Reranker; Right: SapBERT+Reranker}
  \label{fig:EL_reranker_perf}
\end{figure*}

Figure \ref{fig:EL_reranker_perf} shows the final performance of the reranker trained on the candidate generators across three different ablation scenarios—In Dataset, In Domain, and Overall—is presented here. Performance was measured by calculating the macro-average of all metrics (Accuracy@k and MRR) for each semantic category's mentions. We consistently observe that the Overall performance surpasses both in-domain and in-dataset performance across various metrics and semantic types. These results provide strong evidence for the advantages of using a consolidated, canonicalized, uniform resource, such as \dataset{}, to enhance semantic richness and retrieval performance in Biomedical NER and EL. 

\section{Vocabularies Glossary}
Table \ref{tab:vocabularies} presents the top 25 most frequent \biokgs{} in \dataset{}, based on the unique CUIs that map to them. The table provides information on which of these \biokgs{} possess a hierarchical structure and whether their native hierarchy can be accessed through an API or download. As indicated by the highlighted rows, only 11 of these biokgs meet our criteria for extracting full hierarchical paths.
\label{sec:vocabularies_appendix}
\begin{table*}[t!]
\centering
\small
\renewcommand{\arraystretch}{1.10} 

\caption{A comparative overview of biomedical and clinical vocabularies with full names. Counts and percentages are based on the final provided distribution across 41,619 unique CUIs. Vocabularies highlighted in green (prefixed with \textbf{†}) were used to extract hierarchical paths. \textbf{H.} = vocabulary has an internal Hierarchy; \textbf{H.A.} = Hierarchy Available/Accessible for path extraction..}
\label{tab:vocabularies}

\begin{tabular}{
  p{4.8cm}
  S[table-format=4.0]
  S[table-format=2, table-space-text-post={\%}]
  c
  c
  p{4.8cm}
}
\toprule
\textbf{Vocabulary} & {\textbf{Unique CUIs}} & {\textbf{\%}} & \textbf{H.} & \textbf{H.A.} & \textbf{Notes} \\
\midrule
\rowcolor{green!15}
\textbf{† SNOMED CT} (Systematized Nomenclature of Medicine Clinical Terms) & 23261 & 55.89 & \cmark & \cmark & Intl. clinical terminology. Requires free license for full use. \\
\textbf{CHV} (Consumer Health Vocabulary) & 18856 & 45.31 & \xmark & \xmark & Maps consumer terms to professional terms. No standalone API. \\
\rowcolor{green!15}
\textbf{† NCI} (National Cancer Institute Thesaurus) & 16668 & 40.05 & \cmark & \cmark & Comprehensive cancer ontology. Accessible via NCI's EVS and BioPortal. \\
\rowcolor{green!15}
\textbf{† MESH} (Medical Subject Headings) & 13812 & 33.19 & \cmark & \cmark & Thesaurus for indexing literature. API and bulk data from NLM. \\
\textbf{RCD} (Read Codes) & 13435 & 32.28 & \cmark & \xmark & UK primary care codes. Replaced by SNOMED CT. \\
\textbf{SNMI} (SNOMED International) & 11200 & 26.91 & \cmark & \xmark & SNOMED Intl. v3.5. Superseded by SNOMED CT. \\
\rowcolor{green!15}
\textbf{† MDR} (Medical Dictionary for Regulatory Activities) & 7699 & 18.50 & \cmark & \xmark & For adverse events reporting. Requires license. \\
\rowcolor{green!15}
\textbf{† LOINC} (Logical Observation Identifiers Names and Codes) & 6968 & 16.74 & \cmark & \cmark & Laboratory/observation codes. Free with registration. \\
\textbf{SNM} (Systematized Nomenclature of Medicine 1982) & 5926 & 14.24 & \cmark & \xmark & Obsolete SNOMED edition. Replaced by SNOMED CT. \\
\rowcolor{green!15}
\textbf{† LCH\_NW} (Library of Congress Headings, NW Subset) & 5809 & 13.96 & \cmark & \cmark & Northwestern Univ. subset for biomedical topics. \\
\textbf{MEDCIN} & 5653 & {---} & \cmark & \xmark & Proprietary clinical terminology (Medicomp Systems). \\
\textbf{CSP} (CRISP Thesaurus) & 4868 & 11.70 & \cmark & \xmark & Former NIH thesaurus. Now historical; available in UMLS. \\
\textbf{OMIM} (Online Mendelian Inheritance in Man) & 3694 & 8.88 & \xmark & \cmark & No inherent taxonomy. API requires free registration. \\
\textbf{LCH} (Library of Congress Headings) & 3685 & 8.85 & \cmark & \cmark & Broad multidisciplinary subject headings. \\
\textbf{CCPSS} (Canonical Clinical Problem Statement System) & 3662 & 8.80 & \cmark & \xmark & Vanderbilt, 1999. Standard problem names; available via UMLS. \\
\textbf{PSY} (Thesaurus of Psychological Index Terms) & 3137 & 7.54 & \cmark & \xmark & APA's thesaurus for PsycINFO indexing. \\
\rowcolor{green!15}
\textbf{† HPO} (Human Phenotype Ontology) & 2272 & 5.46 & \cmark & \cmark & Open ontology for genetic phenotype annotation. \\
\textbf{FMA} (Foundational Model of Anatomy) & 2192 & 5.27 & \cmark & \cmark & Extensive anatomy ontology with part-of hierarchy. \\
\rowcolor{green!15}
\textbf{† ICD-10-CM} (Intl. Classification of Diseases, 10th Rev, Clin. Mod.) & 2092 & 5.03 & \cmark & \cmark & Clinical mod. of WHO's ICD-10. Maintained by CDC/NCHS. \\
\textbf{RXNORM} & 2042 & 4.91 & \cmark & \xmark & Normalized drug nomenclature. \\
\rowcolor{green!15}
\textbf{† ICD-9-CM} (Intl. Classification of Diseases, 9th Rev, Clin. Mod.) & 1796 & 4.32 & \cmark & \cmark & Legacy system, replaced by ICD-10-CM. \\
\textbf{ICPC2ICD10ENG} (ICPC-2 to ICD-10 Mapping) & 1726 & 3.96 & \xmark & \xmark & Links primary care codes (ICPC-2) to ICD-10. \\
\textbf{UWDA} (Univ. of Washington Digital Anatomist) & 1608 & 3.86 & \cmark & \xmark & Early anatomical ontology; superseded by FMA. \\
\rowcolor{green!15}
\textbf{† GO} (Gene Ontology) & 1378 & 3.31 & \cmark & \cmark & Biological ontology (MF, BP, CC). Open access. \\
\rowcolor{green!15}
\textbf{† NCBI} (National Center for Biotechnology Information) & 1354 & 3.25 & \cmark & \cmark & Suite of databases and tools via E-utilities API. \\
\bottomrule
\end{tabular}
\end{table*}

\end{document}